\documentclass[sigconf, screen]{acmart}
\AtBeginDocument{%
  \providecommand\BibTeX{{%
    \normalfont B\kern-0.5em{\scshape i\kern-0.25em b}\kern-0.8em\TeX}}}

\acmSubmissionID{269}

\usepackage{epsfig}
\usepackage{graphicx}
\usepackage{amsmath}

\usepackage{amssymb}

\usepackage{algorithm}
\usepackage{algorithmic}
\usepackage{multirow}
\usepackage{graphicx}
\usepackage{tikz}
\usepackage{caption}
\usepackage{comment}
\usepackage{amsmath}
\usepackage{amssymb}
\definecolor{citecolor}{RGB}{119,185,0} 

\usepackage{pifont}
\usepackage{color}
\usepackage{colortbl}

\newlength\savewidth\newcommand\shline{\noalign{\global\savewidth\arrayrulewidth
  \global\arrayrulewidth 1pt}\hline\noalign{\global\arrayrulewidth\savewidth}}

\def\eg{\emph{e.g.}} 
\def\ie{\emph{i.e.}} 
\def\etal{\emph{et~al.}} 

\copyrightyear{2024}
\acmYear{2024}
\setcopyright{acmlicensed}
\acmConference[MM '24] {Proceedings of the 32nd ACM International Conference on Multimedia}{October 28--November 1, 2024}{Melbourne, VIC, Australia.}
\acmBooktitle{Proceedings of the 32nd ACM International Conference on Multimedia (MM '24), October 28--November 1, 2024, Melbourne, VIC, Australia}
\acmISBN{979-8-4007-0686-8/24/10}
\acmDOI{10.1145/3664647.3681041}

\begin{document}
\title{Transferring to Real-World Layouts: \\ A Depth-aware Framework for Scene Adaptation}

\author{  
Mu Chen } 
\affiliation{%
  \institution{ReLER Lab, AAII, \\University of Technology Sydney}}
\email{mu.chen@student.uts.edu.au}
\author{  
 Zhedong Zheng}
 \affiliation{%
  \institution{FST and ICI, \\University of Macau}}
\email{zhedongzheng@um.edu.mo}
\author{Yi Yang$^\dagger$} 
 \affiliation{%
  \institution{ReLER Lab, AAII, \\University of Technology Sydney }}
\email{yi.yang@uts.edu.au}

\begin{abstract}
Scene segmentation via unsupervised domain adaptation (UDA) enables the transfer of knowledge acquired from source synthetic data to real-world target data, which largely reduces the need for manual pixel-level annotations in the target domain.
To facilitate domain-invariant feature learning, existing methods typically mix data from both the source domain and target domain by simply copying and pasting pixels.
Such vanilla methods are usually sub-optimal since they do not take into account how well the mixed layouts correspond to real-world scenarios. 
Real-world scenarios are with an inherent layout. 
Real-world scenarios are with an inherent layout.
We observe that semantic categories, such as sidewalks, buildings, and sky, display relatively consistent depth distributions, and could be clearly distinguished in a depth map. 
The model suffers from confusion in predicting the target domain due to the unrealistic mixing. For instance, it is not reasonable to directly paste the near ``pedestrian'' pixels into the remote ``sky'' area.
Based on such observation, we propose a depth-aware framework to explicitly leverage depth estimation to mix categories and facilitate two complementary tasks, \ie, segmentation and depth learning in an end-to-end manner. 
In particular, the framework contains a Depth-guided Contextual Filter (DCF) for data augmentation and a cross-task encoder for contextual learning. 
DCF simulates the real-world layouts, while the cross-task encoder further adaptively fuses the complementing features between two tasks.
Besides, several public datasets do not provide depth annotation. 
Therefore, we leverage the off-the-shelf depth estimation network to obtain the pseudo depth. Extensive experiments show that our methods, even with pseudo depth, achieve competitive performance, 
\ie, 77.7 mIoU on GTA$\rightarrow$Cityscapes and 69.3 mIoU on Synthia$\rightarrow$Cityscapes.

\end{abstract}

\ccsdesc[500]{Computing methodologies~Scene understanding}
\ccsdesc[500]{Computing methodologies~Transfer Learning}

\keywords{Unsupervised Scene Adaptation, Depth Fusion, Transfer Learning}

\maketitle

\begin{figure}[]
  \centering
  \begin{tikzpicture}
    \node[anchor=south west,inner sep=0] (image) at (0,0) {\includegraphics[width=1\linewidth]{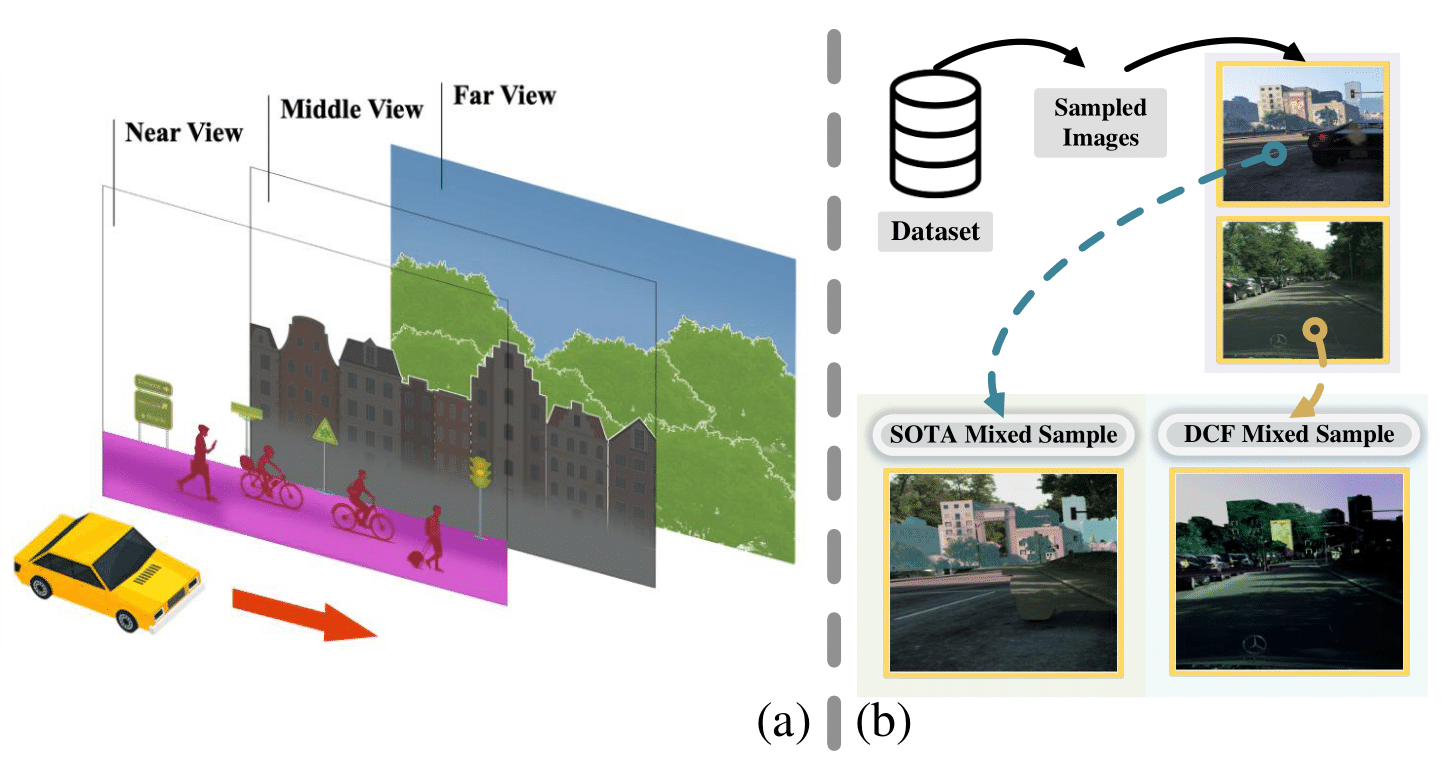}};
    \begin{scope}[x={(image.south east)},y={(image.north west)}]
        \draw[red, thick, dashed] (0.63,0.23) rectangle (0.76,0.36); 
    \end{scope}
  \end{tikzpicture}
  \caption{(a) Considering the driving scenario, we observe that the object location is relatively stable according to the distance from the camera. With such insight, we propose a Depth-guided Contextual Filter (DCF) which is aware of the semantic categories distribution in terms of Near, Middle, and Far view to facilitate cross-domain mixing. 
  (b) Since we explicitly take the semantic layout into consideration, our method achieves more realistic mixed samples compared to existing state-of-the-art methods (Vanilla Mixed Sample)~\cite{chen2023pipa,hoyer2022daformer}. As shown in the \textcolor{red}{red} box, ``new'' buildings are pasted before the parked cars.
  }
  \label{fig1}
\end{figure}

\renewcommand{\thefootnote}{\dag}
\footnotetext[1]{Corresponding author: yi.yang@uts.edu.au}

\section{Introduction}

Semantic segmentation serves as a essential task in machine vision~\cite{wang2024visual,zhou2024prototype,ding2024clustering,li2023semantic,liang2023clustseg,liang2022gmmseg,li2023logicseg,sun2020mining}, benefiting numerous vision applications~\cite{yang2024doraemongpt,chen2024uahoi,yang2024dgl,cheng2023segment,zhou2020target,wang2019zero,zhang2023boosting}. It has achieved significant progress in the last few years \cite{chen2017deeplab, long2015fully, badrinarayanan2017segnet, xie2021segformer, chen2024general,li2022deep,li2023semantic,li2023logicseg,cheng2023segment}. It is worth noting that prevailing models usually require large-scale training datasets with high-quality annotations, such as ADE20K \cite{zhou2019semantic}, to achieve good performance and but such pixel-level annotations in real-world are usually unaffordable and time-consuming \cite{MariusCordts2016TheCD}. 
One straightforward idea is to train networks with synthetic data so that the pixel-level annotations are easier to obtain \cite{StephanRRichter2016PlayingFD, ros2016synthia}. 
However, the network trained with synthetic data results in poor scalability 
when being deployed to a real-world environment due to 
multiple factors, such as weather, illumination, and road design. 
Therefore, researchers resort to unsupervised domain adaptation (UDA) to further tackle the variance between domains~\cite{wang2024disentangled,wang2023disentangled}. 
One branch of UDA methods attempts to mitigate the domain shift by aligning the domain distributions \cite{hoffman2018cycada, ZuxuanWu2019ACEAT, luo2019taking, rangwani2022closer}. Another potential paradigm to heal the domain shift is self-training \cite{YangZou2018UnsupervisedDA, zheng2021rectifying, yang2020fda, li2022class}, which recursively refine the target pseudo-labels. Taking one step further, recent DACS \cite{WilhelmTranheden2020DACSDA} and follow-up works \cite{wang2021domain, hoyer2022daformer, hoyer2022hrda, hoyer2022mic, chen2023pipa, jiang2022prototypical} combine self-training and ClassMix \cite{olsson2021classmix} to 
mix images from both source and target domain. In this way, these works could craft highly perturbed samples to assist training by facilitating learning shared knowledge between two domains. 
Specifically, cross-domain mixing aims to copy the corresponding regions of certain categories from a source domain image and paste them onto an unlabelled target domain image. 
We note that such a vanilla strategy leads to pasting a large amount of objects 
to the unrealistic depth position. 
It is because  
that every category has its own position distribution. 
For instance, background classes such as ``sky'' and ``vegetation'' 
usually appear farther away, while the classes that occupy a small number of pixels such as ``traffic signs'' and ``pole'', usually appear closer as shown in Figure \ref{fig1} (a). 
Such crafted training data compromise contextual learning, leading to 
sub-optimal location prediction performance, especially for small objects. 


To address these limitations, 
we observe the real-world depth distribution and find that  semantic categories are easily separated (disentangled) in the depth map since they follow a similar distribution under certain scenarios, \eg, urban. Therefore, we propose a new depth-aware framework, which contains Depth Contextual Filter (DCF) and a cross-task encoder.  
In particular, DCF removes unrealistic classes mixed with the real-world target training samples based on the depth information.
On the other hand, multi-modal data could improve the performance of deep representations and the effective use of the deep multi-task features to facilitate the final predictions is crucial. The proposed cross-task encoder contains two specific heads to generate intermediate features for each task and an Adaptive Feature Optimization module (AFO). AFO encourages the network to optimize the fused multi-task features in an end-to-end manner. 
Specifically, the proposed AFO adopts a series of transformer blocks to capture the information that is crucial to distinguish different categories and assigns high weights to discriminative features and vice versa. 

The main contributions are as follows: \textbf{(1)} 
We propose a simple Depth-Guided Contextual Filter (DCF) to explicitly leverage the key semantic categories distribution hidden in the depth map, enhancing the realism of cross-domain information mixing and refining the cross-domain layout mixing.
\textbf{(2)} 
We propose an Adaptive Feature Optimization module (AFO) that enables the 
cross-task encoder to exploit the 
discriminative depth information and embed it with 
the visual feature which jointly facilitates semantic segmentation and pseudo depth estimation.
\textbf{(3)} 
Albeit simple, the effectiveness of our proposed methods has been verified by extensive ablation studies. 
Despite the pseudo depth, our method still achieves 
competitive accuracy on two commonly used 
scene adaptation benchmarks, namely 77.7 mIoU on GTA$\rightarrow$Cityscapes and 69.3 mIoU on Synthia$\rightarrow$Cityscapes.

\section{Related Work}
\subsection{Unsupervised Domain Adaptation}
Unsupervised domain adaptation (UDA) aims to train a model on a label-rich source domain and adapt the model to a label-scarce target domain. Some methods propose learning the domain-invariant knowledge by aligning the source and target distribution at different levels. For instance, AdaptSegNet \cite{tsai2018learning}, ADVENT~\cite{vu2019advent}, and CLAN~\cite{luo2019taking} adversarially align the distributions in the feature space. CyCADA~\cite{hoffman2018cycada} diminishes the domain shift at both pixel-level and feature-level representation. DALN~\cite{chen2022reusing} proposes a discriminator-free adversarial learning network and leverages the predicted discriminative information for feature alignment. Both Wu~\etal \cite{ZuxuanWu2019ACEAT} and Yue~\etal~\cite{yue2019domain} learn domain-invariant features by transferring the input images into different styles, such as rainy and foggy, while Zhao~\etal~\cite{zhao2022source} and Zhang~\etal~\cite{zhang2022implicit} diversify the feature distribution via normalization and adding noise respectively. Another line of work refines pseudo-labels gradually under the iterative self-training framework, yielding competitive results. Following the motivation of generating highly reliable pseudo labels for further model optimization, CBST~\cite{YangZou2018UnsupervisedDA} adopts class-specific thresholds on top of self-training to improve the generated labels. Feng~\etal \cite{feng2021complementary} acquire pseudo labels with high precision by leveraging the group information. PyCDA~\cite{lian2019constructing} constructs pseudo-labels in various scales to further improve the training. Zheng~\etal \cite{zheng2019unsupervised} introduce memory regularization to generate consistent pseudo labels. Other works propose either confidence regularization \cite{zou2019confidence, zheng2021rectifying} or category-aware rectification \cite{PanZhang2021PrototypicalPL, zhang2019category} to improve the quality of pseudo labels. DACS \cite{WilhelmTranheden2020DACSDA} proposes a domain-mixed self-training pipeline to mix cross-domain images during training, avoiding training instabilities. Kim~\etal \cite{kim2020learning}, Li~\etal \cite{li2019bidirectional} and Wang~\etal \cite{HaoranWang2020ClassesMA}  combine adversarial and self-training for further improvement. Chen~\etal \cite{chen2022deliberated} establish a deliberated domain bridging (DDB) that aligns and interacts with the source and target domain in the intermediate space. SePiCo~\cite{BinhuiXie2022SePiCoSP} and PiPa~\cite{chen2023pipa} adopt contrastive learning to align the domains. Liu~\etal \cite{liu2022undoing} addresses the label shift problem by adopting class-level feature alignment for conditional distribution alignment. Researchers also attempted to perform entropy minimization \cite{vu2019advent, chen2019domain}, and image translation \cite{ShaohuaGuo2021LabelFreeRC, JinyuYang2020LabelDrivenRF}. consistency regularization\cite{araslanov2021self, choi2019self, melas2021pixmatch, zhou2022context}. Recent multi-target domain adaptation methods enable a single model to adapt a labeled source domain to multiple unlabeled target domains~\cite{gholami2020unsupervised, saporta2021multi, lee2022adas}.
 However, the above methods usually ignore the rich multi-modality information, 
which can be easily obtained from the depth and other sensors.

\subsection{Depth Estimation and Multi-task Learning in Semantic Segmentation}
Semantic segmentation and geometric information are shown to be highly correlated \cite{zhang2018joint, xu2018pad, kanakis2020reparameterizing, vandenhende2020mti, standley2020tasks, zhang2019pattern, WangWQZ021,zhang2024self,DeBNet,liu2024high}.  
Recently depth estimation has been increasingly used to improve the learning of semantics within the context of multi-task learning, but the depth information should be exploited more precisely to help the domain adaptation. SPIGAN \cite{lee2018spigan} pioneered the use of geometric information as an additional supervision by regularizing the generator with an auxiliary depth regression task. DADA \cite{vu2019dada} introduces an adversarial training framework based on the fusion of semantic and depth predictions to facilitate the adaptation. GIO-Ada \cite{chen2019learning} leverages the geometric information on both the input level and output level to reduce domain shift. CTRL \cite{saha2021learning} encodes task dependencies between the semantic and depth predictions to capture the cross-task relationships. CorDA \cite{wang2021domain} bridges the domain gap by utilizing self-supervised depth estimation on both domains. Wu \etal~\cite{wuunsupervised} propose to further support semantic segmentation by depth distribution density. 
Our work follows a similar spirit to leverage depth knowledge as auxiliary supervision. It is worth noting that our work is primarily different from existing works in the following two aspects: (1) from the data perspective, we explicitly delineate the depth distribution to refine data augmentation and construct realistic training samples to enhance contextual learning. (2) from the network perspective, our proposed multi-task learning network not only adopts auxiliary supervision for learning more robust deep representations but also facilitates the multi-task feature fusion by iterative deploying of transformer blocks to jointly learn the rich multi-task information for improving the final predictions.

\section{Method} 

\subsection{Problem Formulation}
In a typical Unsupervised Domain Adaptation (UDA) scenario, we have a source domain, denoted $S$, which consists of abundant labeled synthetic data. On the other hand, the target domain, represented by $T$, contains unlabeled real-world data.
For example, we have labeled training samples $\left(\mathbf{x}^S, \mathbf{y}^S, \mathbf{z}^S \sim \mathbf{X}^S, \mathbf{Y}^S, \mathbf{Z}^S\right)$ in the source domain, where $\mathbf{x}^S, \mathbf{y}^S$ are the training image and the corresponding ground truth for semantic segmentation. $\mathbf{z}^S$ is the label for the depth estimation task. Similarly, we have unlabeled target images sampled from target domain data $\left(\mathbf{x}^T, \mathbf{z}^T \sim \mathbf{X}^T,  \mathbf{Z}^T\right)$, where $\mathbf{x}^T$ is the unlabeled sample in the target domain and $\mathbf{z}^T$ is the label for the depth estimation task. Since depth annotation is not supported by common public datasets, we adopt pseudo depth that can be easily generated by the off-the-shelf model \cite{godard2019digging}. 

\begin{algorithm}[t]
	\renewcommand{\algorithmicrequire}{\textbf{Input:}}
	\renewcommand{\algorithmicensure}{\textbf{Output:}}
	\caption{Depth-guided Contextual Filter Algorithm with Cross-Image Mixing and Self Training}
	\label{alg}
	\begin{algorithmic}[1]
        \REQUIRE Source domain: $(\mathbf{x}^S, \mathbf{y}^S, \mathbf{z}^S \sim \mathbf{X}^S, \mathbf{Y}^S, \mathbf{Z}^S)$, Target domain: $(\mathbf{x}^T, \mathbf{z}^T \sim \mathbf{X}^T, \mathbf{Z}^T)$. Semantic network $\mathcal{F}_{\theta}$.
		\STATE Initialize network parameters $\theta$ randomly.
		\FOR{iteration = 1 to $n$}
        \STATE $\hat{\mathbf{y}}^T \leftarrow 
        \mathcal{F}_\theta\left(\mathbf{x}^T\right),$ Generate pseudo label
        \STATE Pre-calculate the density value $\mathbf{p}$ for each class $i$ at each depth interval from the target depth map $\mathbf{z}^T$,
        \STATE $\hat{\mathbf{y}}^{M} \leftarrow  \mathcal{M} \odot \mathbf{y}^S + \left(1-\mathcal{M}\right) \odot \mathbf{\hat{y}}^T ,$    Randomly select $50\%$ categories and copy the category ground truth label from the source image to target pseudo label\\
        ${\mathbf{x}}^{M} \leftarrow  \mathcal{M} \odot \mathbf{x}^S + \left(1-\mathcal{M}\right) \odot \mathbf{{x}}^T ,$   Copy the corresponding category region from the source image to the target image 
        \STATE Re-calculate the density value $\hat{\mathbf{p}}$ after the mixing,\\
        \STATE Calculate the depth density distribution difference before and after mixing,
        \STATE  Filter the category once the difference exceeds the threshold,\\
        \STATE Re-generate the depth-aware binary mask $\mathcal{M}^{DCF},$\\
        \STATE 
        $\hat{\mathbf{y}}^{F} \leftarrow \mathcal{M^{DCF}} \odot \mathbf{y}^S + \left(1-\mathcal{M^{DCF}}\right) \odot \hat{\mathbf{y}}^T, $ Generate the filtered training samples with new DCF mask\\
        $\mathbf{x}^{F} \leftarrow \mathcal{M^{DCF}} \odot \mathbf{x}^S + \left(1-\mathcal{M^{DCF}}\right) \odot \mathbf{x}^T, $ \\
        \STATE Compute predictions \\ $\bar{\mathbf{y}}^S \leftarrow argmax\left(\mathcal{F}_\theta\left(\mathbf{x}^S\right) \right),$ \\ 
        $\bar{\mathbf{y}}^{F} \leftarrow argmax\left(\mathcal{F}_\theta\left(\mathbf{x}^F\right) \right),$ \\
        \STATE Compute loss for the batch:\\ $\ell \leftarrow {\mathcal{L}}\left(\bar{\mathbf{y}}^S, \mathbf{y}^S, \bar{\mathbf{y}}^F, \hat{\mathbf{y}}^F\right).$
        \STATE Compute $\nabla_\theta \ell $ by backpropagation.
        \STATE Perform stochastic gradient descent. 
        \ENDFOR
        \RETURN $\mathcal{F}_{\theta}$
	\end{algorithmic}  
\end{algorithm}

\begin{figure}[]
  \centering
  \includegraphics[width=1\linewidth]{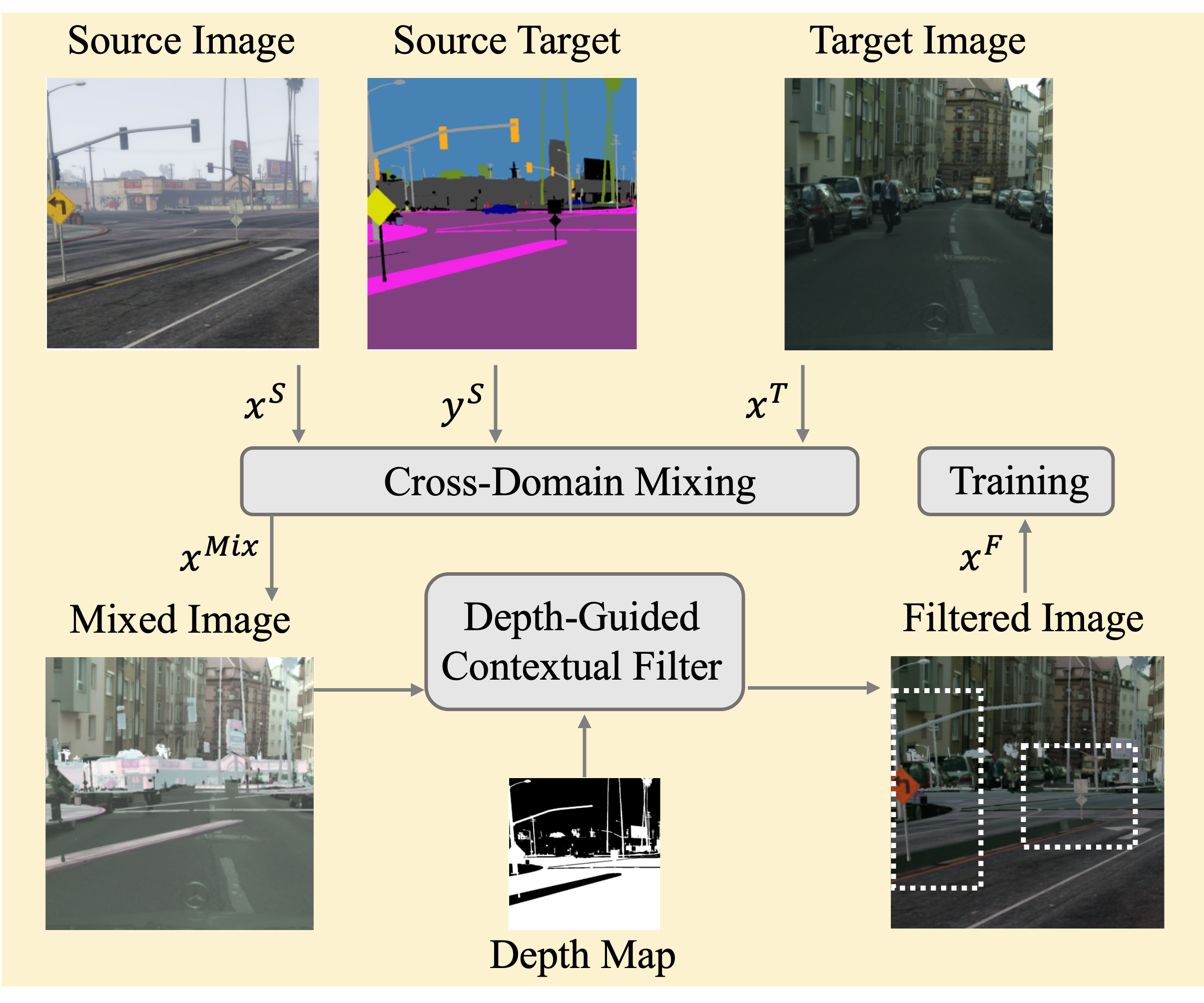}
  \caption{Source domain images $x^S$ and $x^T$ are mixed together, using the ground truth label $y^S$. The mixed images are de-noised by our proposed Depth-guided Contextual Filter (DCF) and then trained by the network. We illustrate DCF with a set of practical sample. As illustrated, the unrealistic ``Building'' pixels from the source image are mixed pasted to the target image, leading to a noisy mixed sample. DCF removes these pixels and maintain mixed pixels of ``Traffic Sign'' and ``Pole'' shown in the white dotted boxes, enhancing the realism of cross-domain mixing. (Best viewed when zooming in.)}
  \label{fig2}
\end{figure} 

\begin{figure*}[]
  \centering
  \includegraphics[width=0.95\linewidth]{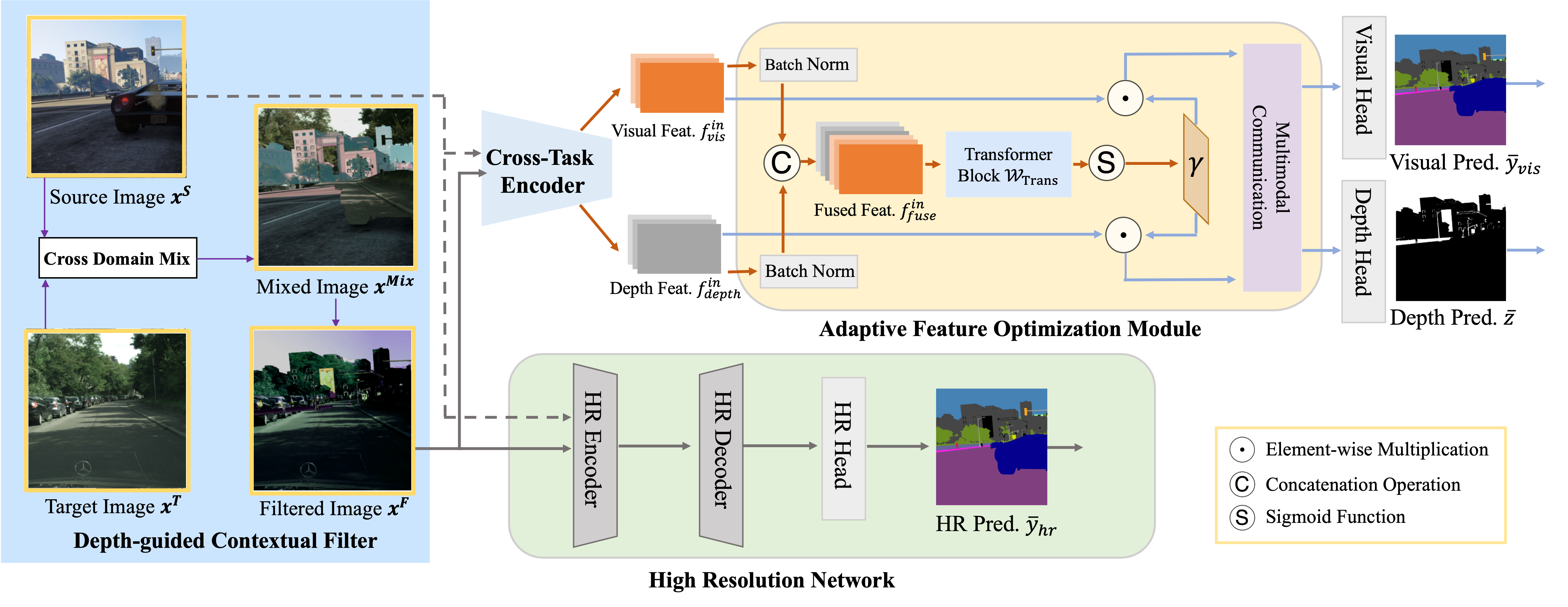}
  \put(-5,155){{\scriptsize $\mathcal{L}_{vis}$}} 
  \put(-5,106){{\scriptsize $\mathcal{L}_{depth}$}} 
  \put(-143,46){{\scriptsize $\mathcal{L}_{hr}$}} 
  \caption{The proposed multi-task learning framework. The input images $x^F$ are mixed from the source image $x^S$ and target domain $x^T$ according to the depth (Please refer to Figure~\ref{fig2}).  
  Then we are fed $x^S$ and $x^F$ into the high resolution encoder to generate high resolution predictions. 
  To enhance multi-modal learning, the visual and depth feature created by the cross-task encoder are fused and fed into the proposed Adaptive Feature Optimization module (AFO) for multimodal communication. Finally, the multimodal communication via several transformer blocks incorporates and optimizes the fusion of depth information, improving the final visual predictions.}
  \label{fig3}
\end{figure*}

\subsection{Depth-guided Contextual Filter}
In UDA, recent works 
Recent UDA works~\cite{olsson2021classmix, wang2021domain, hoyer2022daformer, hoyer2022hrda, hoyer2022mic, chen2023pipa} 
often employ pixel mixing to create cross-domain augmented samples. The basic idea is straightforward: take a portion of pixels from a source domain image and transplant them onto an equivalent area in a target domain image. However, this simple approach faces challenges due to the inherent differences in structure and layout between source and target domain data.
To decrease noisy signals and simulate augmented training samples with real-world layouts, we propose Depth-guided Contextual Filter (DCF) to reduce the noisy pixels that are naively mixed across domains. The implementation of DCF is represented as pseudo-code in Algorithm \ref{alg}, where the image $\mathbf{x}^S$ and the corresponding semantic labels $\mathbf{y}^S$ are sampled from source domain data. The image $\mathbf{x}^T$ and the depth label $\mathbf{z}^T$ are from target domain data. Pseudo label $\hat{\mathbf{y}}^T$ is then generated as $\hat{\mathbf{y}}^T=\mathcal{F}_\theta\left(\mathbf{x}^T\right)$, where $\mathcal{F}_\theta$ is a pre-trained semantic network. In practice, $\mathcal{F}_\theta$ usually has been trained on the source domain dataset via supervised learning.
Based on the hypothesis that most semantic categories usually fall under a finite depth range, we introduce DCF, which divides the target depth map $\mathbf{z}^T$ into a few discrete depth intervals $(\Delta z_1,..., \Delta z_n)$. 
For a given real-world target input image $\mathbf{x}^T$ combined with the pseudo label $\hat{\mathbf{y}}^T$ and target depth map $\mathbf{z}^T$, the density value at each depth interval $(\Delta z_1,..., \Delta z_n)$ for each class $i\in(1,\ldots, C)$ can be 
counted and normalized as a probability.
We denote the density value for class $i$ at the depth interval $\Delta z_1$ as $p_i(\Delta z_1)$. 
All the density values make up the depth distribution in the target domain image. 
Then we randomly select half of the categories on the source images to paste on the target domain image. 
In practice, we apply a binary mask $\mathcal{M}$ to denote the corresponding pixels. 
Then naive cross-domain mixed image $\mathbf{x}^{Mix}$ and the mixed label $\hat{\mathbf{y}}^{Mix}$ 
can be formulated as:
\begin{align}
\mathbf{x}^{Mix}=\mathcal{M} \odot \mathbf{x}^S + \left(1-\mathcal{M}\right) \odot \mathbf{x}^T, \\
\hat{\mathbf{y}}^{Mix}=\mathcal{M} \odot \mathbf{y}^S + \left(1-\mathcal{M}\right) \odot \hat{\mathbf{y}}^T,
\end{align}
where $\odot$ denotes the element-wise multiplication of between the mask and the image. The naively mixed images are visualized in Figure \ref{fig2}. It could be observed that due to the depth distribution difference between two domains, pixels of ``Building'' category are mixed from the source domain to the target domain, creating unrealistic images. Training with such training samples will compromise contextual learning. Therefore, we propose to filter the pixels that do not match the depth density distribution in the mixed image. After the naive mixing, we re-calculate the density value for each class at each depth interval. For example, the new density value for class $i$ at the depth interval $\Delta z_1$ is denoted as $\hat{p}_i\left(\Delta z_1\right)$. Then we calculate the depth density distribution difference for each pasted category and denote the difference for class $i$ at the depth interval $\Delta z_1$ as $\Delta p_i(\Delta z_1) = |p_i(\Delta z_1) - \hat{p}_i(\Delta z_1)|$. Once $\Delta p_i(\Delta z_1)$ exceeds the threshold of that category $i$, these pasted pixels are removed.
After performing DCF, we confirm the final realistic pixels to be mixed and construct a depth-aware binary mask $\mathcal{M^{DCF}}$, which is changed dynamically based on the depth layout of the current target image. 

The filtered mixing samples are then generated. In practice, we directly apply the updated depth-aware mask to replace the original mask. Therefore, the new mixed sample and the label are as follows:
\begin{align}
\mathbf{x}^{F}=\mathcal{M^{DCF}} \odot \mathbf{x}^S + \left(1-\mathcal{M^{DCF}}\right) \odot \mathbf{x}^T, \\
\hat{\mathbf{y}}^{F}=\mathcal{M^{DCF}} \odot \mathbf{y}^S + \left(1-\mathcal{M^{DCF}}\right) \odot \hat{\mathbf{y}}^T.
\end{align}
Because large objects such as ``sky'' and ``terrain'' usually aggregate and occupy a large amount of pixels and small objects only occupy a small amount of pixels in a certain depth range, we set different filtering thresholds for each category. DCF uses pseudo semantic labels for the target domain as there is no ground truth available. Since the label prediction is not stable in the early stage, we apply a warmup strategy to perform DCF after 10,000 iterations. Examples of the input images, naively mixed samples and filtered samples are presented in Figure \ref{fig2}. The sample after the process of the DCF module has the pixels from the source domain that match the depth distribution of the target domain, helping the network to better deal with the domain gap.

\subsection{Multi-task Scene Adaptation Framework}
To exploit the relation between segmentation and depth learning, we introduce a multi-task scene adaptation framework including a high resolution semantic encoder, and a cross-task shared encoder with a feature optimization module, which is depicted in Figure \ref{fig3}. The proposed framework incorporates and optimizes the fusion of depth information for improving the final semantic predictions.

\paragraph{High Resolution Semantic Prediction.}
Most supervised methods use high resolution images for training, but common scene adaptation methods usually use random crops of the image that is half of the full resolution. To reduce the domain gap between scene adaptation and supervised learning while maintaining the GPU memory consumption, we adopt a high-resolution encoder to encode HR image crops into deep HR features. Then a semantic decoder is used to generate the HR semantic predictions $\bar{\mathbf{y}}_{hr}$. We adopt the cross entropy loss for semantic segmentation: 
\begin{align}
{\mathcal{L}}_{hr}^S=\mathbb{E}\left[- \mathbf{y}^S \log \bar{\mathbf{y}}_{hr}^S\right], \quad 
{\mathcal{L}}_{hr}^F=\mathbb{E}\left[- \hat{\mathbf{y}}^F \log \bar{\mathbf{y}}_{hr}^F\right],
\end{align}
where $\bar{\mathbf{y}}_{hr}^S$ and $\bar{\mathbf{y}}_{hr}^T$ are high resolution semantic predictions. $\mathbf{y}^S$ is the one-hot semantic label for the source domain and $\hat{\mathbf{y}}^F$ is the one-hot pseudo label for the depth-aware fused domain. 

\paragraph{Adaptive Feature Optimization.}
In addition to the high resolution encoder, We use another cross-task encoder to encode input images which are shared for both tasks. Depth maps are rich in spatial depth information, but a naive concatenation of depth information directly to visual information causes some interference, e.g. categories at similar depth positions are already well distinguished by visual information, and attention mechanisms can help the network to select the crucial part of the multitask information. In the proposed multi-task learning framework, the visual semantic feature and depth feature is  generated by a visual head and a depth head, respectively. As shown in Figure~\ref{fig3}, after applying batch normalization, an Adaptive Feature Optimization module then concatenates the normalized input visual feature and the input depth feature to create a fused multi-task feature by concatenation as 
$f_{fuse}^{in} = \operatorname{CONCAT}\left(f_{vis}^{in} , f_{depth}^{in}\right).$
The fused feature is then fed into a series of transformer blocks to capture the key information between the two tasks. The attention mechanism adaptively adjusts the extent to which depth features are embedded in visual features:
\begin{equation}
f_{fuse}^{out}= \mathcal{W}_{Trans}\left(f_{fuse}^{in}\right),
\end{equation}
where $\mathcal{W}_{Trans}$ is the transformer parameter.
The learned output of the transformer blocks is a weight map $\gamma$ which is multiplied back to the input visual feature and depth feature resulting in an optimized feature as: 
\begin{equation}
\boldsymbol{\gamma}=\boldsymbol{\sigma}\left(\mathcal{W}_{Conv} \otimes f_{fuse}^{out}\right),
\end{equation}
where $\mathcal{W}_{Conv}$ denotes the convolution parameter, $\otimes$ denotes the convolution operation and $\boldsymbol{\sigma}$ represents the sigmoid function.
The weight matrix $\boldsymbol{\gamma}$ performs adaptive optimization of the muti-task features. Then, the fused feature $f_{fuse}^{out}$ is fed into different decoders for predicting different final tasks, \ie, the visual and the depth task.
The output features are essentially multimodal features containing crucial depth information:
\begin{equation}
f_{vis}^{out} = f_{vis}^{in} \odot \boldsymbol{\gamma}, \quad
f_{depth}^{out} = f_{depth}^{in} \odot \boldsymbol{\gamma},
\end{equation}
where $\odot$ represents element-wise multiplication.
The optimized visual and depth feature is then fed into the multimodal communication module for further processing. The multimodal communication module refines the learning of key information between two tasks by iterative use of transformer blocks. the inference is merely based on the visual input when the feature optimization is fished. The final semantic prediction $\bar{\mathbf{y}}_{vis}^S$ and depth prediction $\bar{\mathbf{z}}^S$  can be generated from the final visual feature $f_{vis}^{final}$ and depth feature $f_{depth}^{final}$ by visual head and depth head. Similar to high resolution predictions, we use the cross entropy loss for the semantic loss calculation:
\begin{equation}
{\mathcal{L}}_{vis}^S=\mathbb{E}\left[- \mathbf{y}^S \log \bar{\mathbf{y}}_{vis}^S\right],\quad 
{\mathcal{L}}_{vis}^F=\mathbb{E}\left[-\hat{\mathbf{y}}^F \log \bar{\mathbf{y}}_{vis}^F \right].
\end{equation}
We also employ berHu loss for depth regression at source domain:
\begin{equation}
{\mathcal{L}}_{depth}^S= \mathbb{E}\left[\operatorname{berHu}\left(\bar{\mathbf{z}}^S-\mathbf{z}^S\right)\right],
\end{equation}
where $\bar{z}$ and $z$ are predicted and ground truth semantic maps. Following \cite{vu2019dada, saha2021learning}, we deploy the reversed Huber criterion~\cite{laina2016deeper}, which is defined as :
\begin{equation}
\begin{aligned}
\operatorname{ber}Hu\left(e_z\right) & =\left\{
\begin{array}{cc}
\left|e_z\right|, & \left|e_z\right| \leq H \\
\frac{\left(e_z\right)^2+H^2}{2H}, & \left|e_z\right|>H
\end{array}
\right. \\
H & =0.2 \max \left(\left|e_z\right|\right),
\end{aligned}
\end{equation}
where $H$ is a positive threshold and we set it to $0.2$ of the maximum depth residual.
Finally, the overall loss function is:
\begin{equation}
{\mathcal{L}} = {\mathcal{L}}_{hr}^S + {\mathcal{L}}_{vis}^S  + \lambda_{depth}{\mathcal{L}}_{depth}^S + {\mathcal{L}}_{hr}^F + {\mathcal{L}}_{vis}^F,
\end{equation}
where hyperparameter 
$\lambda_{depth}$ is the loss weight. Considering that our main task is semantic segmentation and the depth estimation is the auxiliary task, we empirically $\lambda_{depth}$ = $\lambda_{depth}$ = 1 $\times 10^{-3}$. We also designed the ablation studies to change the weight of depth task $\lambda_{depth}$ to the level of $10^{-1}$ or $10^{-3}$. 

\begin{figure*}[]
  \centering
  \includegraphics[width=0.95\linewidth]{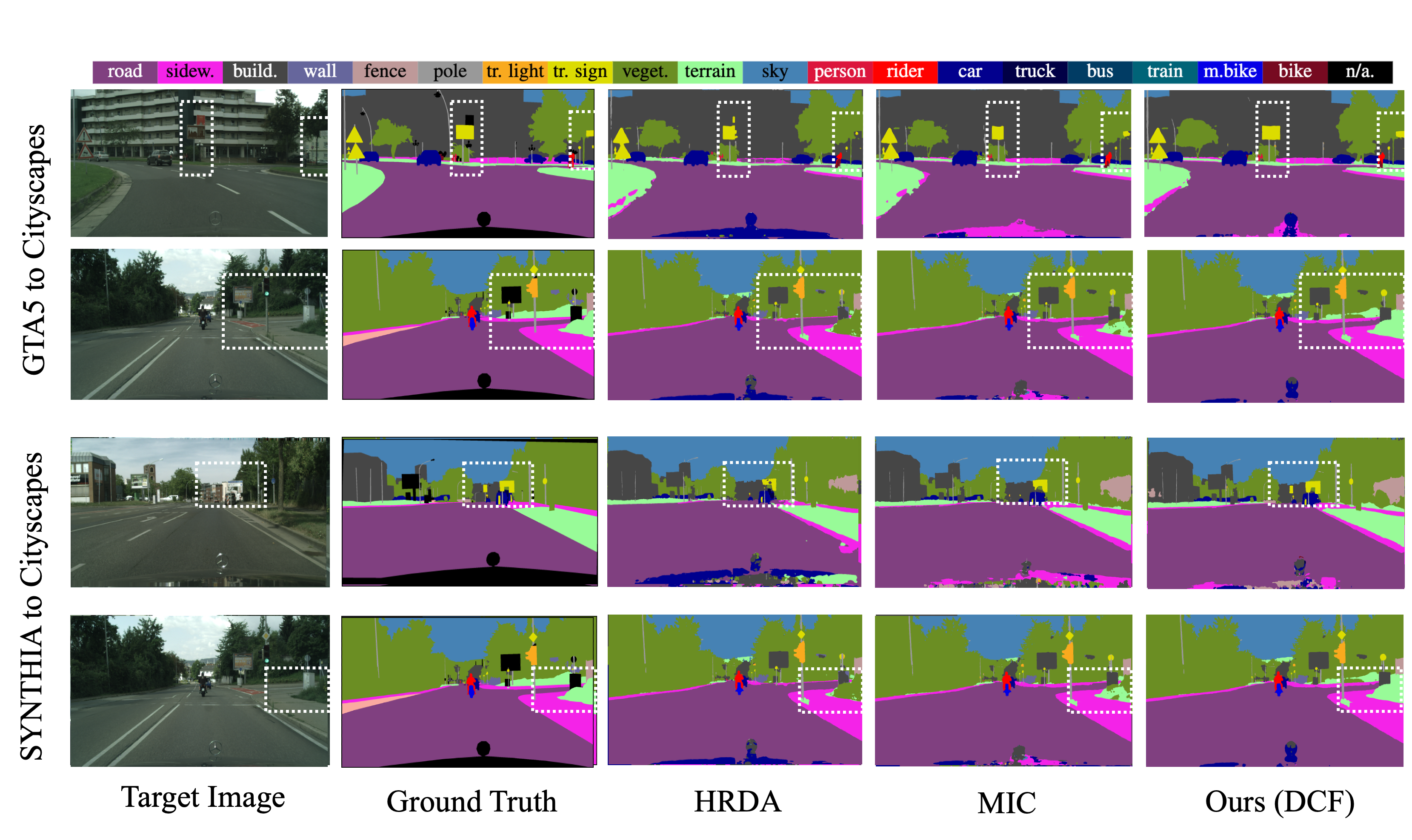}
  \caption{Qualitative results. From left to right: Target Image, Ground Truth, the visual results predicted by HRDA, MIC and Ours. We highlight prediction differences in white dash boxes and it is observed that the proposed method predicts clear edges.}
  \label{fig4}
\end{figure*}

\section{Experiment} 

\subsection{Implementation Details}
\noindent\textbf{Datasets.}
We evaluate the proposed framework on two scene adaptation settings, \ie, GTA $\rightarrow$ Cityscapes and SYNTHIA $\rightarrow$ Cityscapes, following common protocols \cite{WilhelmTranheden2020DACSDA, wang2021domain, araslanov2021self, hoyer2022daformer, hoyer2022hrda, hoyer2022mic}. Particularly, the GTA5 dataset~\cite{StephanRRichter2016PlayingFD} is the synthetic dataset collected from a video game, which contains 24,966 images annotated by 19 classes. Following \cite{wang2021domain}, we adopt depth information generated by Monodepth2~\cite{godard2019digging} model which is trained merely on GTA image sequences. SYNTHIA~\cite{ros2016synthia} is a synthetic urban scene dataset with 9,400 training images and 16 classes. Simulated depth information provided by SYNTHIA is adopted. GTA and SYNTHIA serve as source domain datasets. The target domain dataset is Cityscapes, which is collected from real-world street-view images. Cityscapes contains 2,975 unlabeled training images and 
500 validation images. The resolution of Cityscapes is 2048  $\times$ 1024 and the common protocol downscales the size to 1024 $\times$ 512 to save memory. Following \cite{wang2021domain}, the stereo depth estimation from \cite{sakaridis2018model} is used. We leverage the Intersection Over Union (IoU) for per-class performance and the mean Intersection over Union (mIoU) over all classes to report the result. The code is based on Pytorch~\cite{paszke2017automatic}.

\noindent\textbf{Experimental Setup.}
We adopt DAFormer \cite{hoyer2022daformer} network with MiT-B5 backbone~\cite{xie2021segformer} for the high resolution encoder and DeepLabV2 network with ResNet-101 backbone for the cross-task encoder to reduce the memory consumption. All backbones are initialized with ImageNet pretraining. Our training procedure is based on self-training methods with cross-domain mixing \cite{WilhelmTranheden2020DACSDA, hoyer2022daformer, hoyer2022hrda, hoyer2022mic} and enhanced by our proposed Depth-guided Contextual Filter. Following \cite{WilhelmTranheden2020DACSDA, hoyer2022hrda}, the input image resolution is half of the full resolution for the cross-task encoder and full resolution for high resolution encoder. We utilize the same data augmentation, \eg, color jitter and Gaussian blur and empirically set pseudo labels threshold $0.968$ following \cite{WilhelmTranheden2020DACSDA}. We train the network with batch size 2 for 40k iterations on a Tesla V100 GPU. 

\noindent\textbf{Data Resolution.}
Our proposed depth-aware multi-task framework contains a high resolution encoder and a cross-task encoder with an adaptive feature optimization module (AFO). Previous works~\cite{WilhelmTranheden2020DACSDA, tsai2019domain, li2019bidirectional} downsample Cityscapes to 1024 $\times$ and GTA to 1280 $\times$ 720. Following \cite{hoyer2022hrda}, for the high resolution encoder, we resize GTA to 2560 $\times$ 1440 and SYNTHIA to 2560 x 1520. Then the crop size is 1024 $\times$ 1024. In addition,  SegFormer~\cite{xie2021segformer} MLP decoder with an embedding dimension of 256 is used for the high resolution branch. For the cross-task encoder branch, we follow common UDA methods~\cite{WilhelmTranheden2020DACSDA, hoyer2022daformer} to adopt 1024 $\times$ 512 pixels (half of the full resolution) for Cityscapes, 1280 $\times$ 760 for SYNTHIA and 1280 $\times$ 720 for GTA. In addition, a 512 $\times$ 512 random crop is extracted. 

\subsection{Comparison with SOTA}

\begin{table*}[!t]
	\centering
	\caption{Quantitative comparison with previous UDA methods on GTA $\rightarrow$ Cityscapes. We present pre-class IoU and mIoU. The best accuracy in every column is in \textbf{bold}. Our results are averaged over 3 random seeds.
 }
	\label{table:gtacity} 
	\resizebox{\linewidth}{!}{
	\begin{tabular}{c|ccccccccccccccccccc|c}
		\shline
		Method & Road & SW & Build & Wall & Fence & Pole & TL & TS & Veg. & Terrain & Sky & PR & Rider & Car & Truck & Bus & Train & Motor & Bike & mIoU\\
		\shline
		\hline
		AdaptSegNet~\cite{tsai2018learning} & 86.5 & 36.0 & 79.9& 23.4 & 23.3 & 23.9 & 35.2 & 14.8 & 83.4 & 33.3 & 75.6 & 58.5 & 27.6 & 73.7 & 32.5 & 35.4 & 3.9 & 30.1 & 28.1 & 42.4\\ 
		CyCADA \cite{hoffman2018cycada} & 86.7 & 35.6 & 80.1 & 19.8 & 17.5 & 38.0 & 39.9 & 41.5 & 82.7 & 27.9 & 73.6 & 64.9 & 19.0 & 65.0 & 12.0 & 28.6 & 4.5 & 31.1 & 42.0 & 42.7 \\
		CLAN~\cite{luo2019taking} & 87.0 & 27.1 & 79.6 & 27.3 & 23.3 & 28.3 & 35.5 & 24.2 & 83.6 & 27.4 & 74.2 & 58.6 & 28.0 & 76.2 & 33.1 & 36.7 & 6.7 & 31.9 & 31.4 & 43.2 \\
		SP-Adv~\cite{SHAN2020125} & 86.2 & 38.4 & 80.8 & 25.5& 20.5 & 32.8 & 33.4 & 28.2 & 85.5 & 36.1 & 80.2 & 60.3 & 28.6 & 78.7 & 27.3 & 36.1 &	4.6	& 31.6 & 28.4 & 44.3 \\
		MaxSquare~\cite{chen2019domain} & 88.1 & 27.7 & 80.8 & 28.7 & 19.8 & 24.9 & 34.0 & 17.8 & 83.6 & 34.7 & 76.0 & 58.6 & 28.6 & 84.1 & 37.8 & 43.1 & 7.2 & 32.3 & 34.2 & 44.3 \\
		ASA~\cite{zhou2020affinity} & 89.2 & 27.8 & 81.3 & 25.3 & 22.7 & 28.7 & 36.5 & 19.6 & 83.8 & 31.4 & 77.1 & 59.2 & 29.8 & 84.3 & 33.2 & 45.6 & 16.9 & 34.5 & 30.8 & 45.1 \\
		AdvEnt~\cite{vu2019advent} & 89.4 & 33.1 & 81.0 & 26.6 & 26.8 & 27.2 & 33.5 & 24.7 & 83.9 & 36.7 & 78.8 & 58.7 & 30.5 & 84.8 & 38.5 & 44.5 & 1.7 & 31.6 & 32.4 & 45.5 \\
		MRNet~\cite{zheng2019unsupervised}  & 89.1 & 23.9 & 82.2 & 19.5 & 20.1 & 33.5 & 42.2 & 39.1 & 85.3 & 33.7 & 76.4 & 60.2 & 33.7 & 86.0 & 36.1 & 43.3 & 5.9 & 22.8 & 30.8 & 45.5 \\
		APODA~\cite{yang2020adversarial}  & 85.6 & 32.8 & 79.0 & 29.5 & 25.5 & 26.8 & 34.6 & 19.9 & 83.7 & 40.6 & 77.9 & 59.2 & 28.3 & 84.6 & 34.6 & 49.2 & 8.0 & 32.6 & 39.6 & 45.9 \\
		CBST \cite{YangZou2018UnsupervisedDA} & 91.8 & 53.5 & 80.5 & 32.7 & 21.0 & 34.0 & 28.9 & 20.4 & 83.9 & 34.2 & 80.9 & 53.1 & 24.0 & 82.7 & 30.3 & 35.9 & 16.0 & 25.9 & 42.8 & 45.9\\
	MRKLD \cite{zou2019confidence} & 91.0 & 55.4 & 80.0 & 33.7 & 21.4 & 37.3 & 32.9 & 24.5 & 85.0 & 34.1 & 80.8 & 57.7 & 24.6 & 84.1 & 27.8 & 30.1 & 26.9 & 26.0 & 42.3 & 47.1\\

        FADA~\cite{HaoranWang2020ClassesMA} & 91.0 & 50.6 & 86.0 & 43.4 & 29.8 & 36.8 & 43.4 & 25.0 & 86.8 & 38.3 & 87.4 & 64.0 & 38.0 & 85.2 & 31.6 & 46.1 & 6.5 & 25.4 & 37.1 & 50.1\\ 
		Uncertainty~\cite{zheng2021rectifying} & 90.4 & 31.2 & 85.1 & 36.9 & 25.6 & 37.5 & 48.8 & 48.5 & 85.3 & 34.8 & 81.1 & 64.4 & 36.8 & 86.3 & 34.9 & 52.2 & 1.7 & 29.0 & 44.6 & 50.3 \\
		FDA~\cite{yang2020fda} & 92.5 & 53.3 & 82.4 & 26.5 & 27.6 & 36.4 & 40.6 & 38.9 & 82.3 & 39.8 & 78.0 & 62.6 & 34.4 & 84.9 & 34.1 & 53.1 & 16.9 & 27.7 & 46.4 & 50.5\\
	    Adaboost~\cite{zheng2022adaptive} & 90.7 & 35.9 & 85.7 & 40.1 & 27.8 & 39.0 & 49.0 & 48.4 & 85.9 & 35.1 & 85.1 & 63.1 & 34.4 & 86.8 & 38.3 & 49.5 & 0.2 & 26.5 & 45.3 & 50.9\\
		DACS \cite{WilhelmTranheden2020DACSDA} & 89.9 & 39.7 & 87.9 & 30.7 & 39.5 & 38.5 & 46.4 & 52.8 & 88.0 & 44.0 & 88.8 & 67.2 & 35.8 & 84.5 & 45.7 & 50.2 & 0.0 & 27.3 & 34.0 & 52.1\\

		BAPA \cite{liu2021bapa} & 94.4 & 61.0 & 88.0 & 26.8 & 39.9 & 38.3 & 46.1 & 55.3 & 87.8 & 46.1 & 89.4 & 68.8 & 40.0 & 90.2 & 60.4 & 59.0 & 0.0 & 45.1 & 54.2 & 57.4\\
		ProDA \cite{PanZhang2021PrototypicalPL} & 87.8 & 56.0 & 79.7 & 46.3 & 44.8 & 45.6 & 53.5 & 53.5 & 88.6 & 45.2 & 82.1 & 70.7 & 39.2 & 88.8 & 45.5 & 59.4 & 1.0 & 48.9 & 56.4 & 57.5\\
        CaCo \cite{huang2022category} & 93.8 & 64.1 & 85.7 & 43.7 & 42.2 & 46.1 & 50.1 & 54.0 & 88.7 & 47.0 & 86.5 & 68.1 & 2.9 & 88.0 & 43.4 & 60.1 & 31.5 & 46.1 & 60.9 & 58.0\\
		DAFormer \cite{hoyer2022daformer} & 95.7 & 70.2 & 89.4 & 53.5 & 48.1 & 49.6 & 55.8 & 59.4 & 89.9 & 47.9 & 92.5 & 72.2 & 44.7 & 92.3 & 74.5 & 78.2 & 65.1 & 55.9 & 61.8 & 68.3\\
        CAMix \cite{zhou2022context} & 96.0 & 73.1 & 89.5 & 53.9 & 50.8 & 51.7 & 58.7 & 64.9 & 90.0 & 51.2 & 92.2 & 71.8 & 44.0 & 92.8 & 78.7 & 82.3 & 70.9 & 54.1 & 64.3 & 70.0\\
		HRDA \cite{hoyer2022hrda} & 96.4 & 74.4 & 91.0 & 61.6 & 51.5 & 57.1 & 63.9 & 69.3 & 91.3 & 48.4 & 94.2 & 79.0 & 52.9 & 93.9 & 84.1 & 85.7 & 75.9 & 63.9 & 67.5 & 73.8\\
		MIC \cite{hoyer2022mic} & 97.4 & 80.1 & 91.7 & 61.2 & 56.9 & 59.7 & 66.0 & 71.3 & 91.7 & 51.4 & \textbf{94.3} & 79.8 & 56.1 & 94.6 & 85.4 & \textbf{90.3} & 80.4 & 64.5 & 68.5 & 75.9\\
                \hline
        CorDA$^\dagger$ \cite{wang2021domain} & 94.7 & 63.1 & 87.6 & 30.7 & 40.6 & 40.2 & 47.8 & 51.6 & 87.6 & 47.0 & 89.7 & 66.7 & 35.9 & 90.2 & 48.9 & 57.5 & 0.0 & 39.8 & 56.0 & 56.6\\
                FAFS$^\dagger$ \cite{cardace2022plugging} & 93.4 & 60.7 & 88.0 & 43.5 & 32.1 & 40.3 & 54.3 & 53.0 & 88.2 & 44.5 & 90.0 & 69.5 & 35.8 & 88.7 & 34.1 & 53.9 & 41.3 & 51.7 & 54.7 & 58.8\\
        DBST$^\dagger$ \cite{cardace2022plugging} & 94.3 & 60.0 & 87.9 & 50.5 & 43.0 & 42.6 & 50.8 & 51.3 & 88.0 & 45.9 & 89.7 & 68.9 & 41.8 & 88.0 & 45.8 & 63.8 & 0.0 & 50.0 & 55.8 & 58.8\\
  		Ours$^\dagger$  & \textbf{97.5} & \textbf{80.7} & \textbf{92.1} & \textbf{66.4} & \textbf{62.3} & \textbf{63.1} & \textbf{67.7} & \textbf{75.7} & \textbf{91.8} & \textbf{52.4} & 93.9 & \textbf{81.6} & \textbf{61.8} & \textbf{94.7} & \textbf{88.3} & 90.0 & \textbf{81.2} & \textbf{65.8} & \textbf{69.6} & \textbf{77.7}\\
        \hline
		\shline
	\end{tabular}
	}
       \noindent\raggedright\footnotesize{$^\dagger$: Training with depth data.} 
\end{table*} 

\noindent\textbf{Results on GTA$\rightarrow$Cityscapes.}
We show our results on GTA $\rightarrow$ Cityscapes in Table \ref{table:gtacity} and highlight the best results in bold. Our method yields significant performance improvement over the state-of-the-art method MIC \cite{hoyer2022mic} from 75.9 mIoU to 77.7 mIoU. Usually, classes that occupy a small number of pixels are difficult to adapt and have a comparably low IoU performance. However, our method demonstrates competitive IoU improvement in most categories especially on small objects such as +5.7 on ``Rider'', +5.4 on ``Fence'', +5.2 on ``Wall'', +4.4 on ``Traffic Sign'' and +3.4 on ``Pole''. The result shows the effectiveness of the proposed contextual filter and cross-task learning framework in contextual learning. Our method also increases the mIoU performance of classes that aggregate and occupy a large amount of pixels in an image by a smaller margin such as +1.8 on ``Pedestrain'' and +1.1 on ``Bike'', probably because the rich texture and color information contained in the visual feature already has the ability to recognize these relatively easier classes. The above observations are also qualitatively reflected in Figure~\ref{fig4}, where we visualize the segmentation results of the proposed method and the comparison with previous strong transformer-based methods HRDA \cite{hoyer2022hrda}, and MIC \cite{hoyer2022mic}. The qualitative results highlighted by white dash boxes show that the proposed method largely improved the prediction quality of challenging small object ``Traffic Sign'' and large category ``Terrain''.

\begin{table*}[!t]
	\centering
	\caption{Quantitative comparison with previous UDA methods on SYNTHIA $\rightarrow$ Cityscapes. We present pre-class IoU, mIoU and mIoU*. mIoU and mIoU* are averaged over 16 and 13 categories, respectively. The best accuracy in every column is in \textbf{bold}.Our results are averaged over 3 random seeds.}
	\label{table:syncity}
	\resizebox{\linewidth}{!}{
	\begin{tabular}{c|cccccccccccccccc|c|c}
		\shline
		Method & Road & SW & Build & Wall* & Fence* & Pole* & TL & TS & Veg. & Sky & PR & Rider & Car & Bus & Motor & Bike & mIoU* & mIoU\\
		\shline
		\hline
		MaxSquare~\cite{chen2019domain} & 77.4 & 34.0 & 78.7  & 5.6 & 0.2 & 27.7 & 5.8 & 9.8 & 80.7 & 83.2 & 58.5 & 20.5 & 74.1 & 32.1 & 11.0 & 29.9 & 45.8 & 39.3 \\
		SIBAN~\cite{luo2019significance} & 82.5 & 24.0 & 79.4 & $-$ & $-$ & $-$ & 16.5 & 12.7 & 79.2 & 82.8 & 58.3 & 18.0 & 79.3 & 25.3 & 17.6 & 25.9 & 46.3 & $-$ \\
    	PatchAlign~\cite{tsai2019domain} & 82.4 & 38.0 & 78.6 & 8.7 & 0.6 & 26.0 & 3.9 & 11.1 & 75.5 & 84.6 & 53.5 & 21.6 & 71.4 & 32.6 & 19.3 & 31.7 & 46.5 & 40.0 \\
		AdaptSegNet~\cite{tsai2018learning} & 84.3 & 42.7 & 77.5 & $-$ & $-$ & $-$ & 4.7 & 7.0 & 77.9 & 82.5 & 54.3 & 21.0 & 72.3 & 32.2 & 18.9 & 32.3 & 46.7 & $-$ \\
		CLAN~\cite{luo2019taking} & 81.3 & 37.0 & 80.1 & $-$ & $-$ & $-$ & 16.1 & 13.7 & 78.2 & 81.5 & 53.4 & 21.2 & 73.0 & 32.9 & 22.6 & 30.7 & 47.8 & $-$  \\
		SP-Adv~\cite{SHAN2020125} & 84.8 & 	35.8 & 	78.6 & $-$ & $-$ & $-$ &	6.2	& 15.6 & 	80.5 &	82.0 &	66.5 & 	22.7 & 	74.3 &	34.1	 & 19.2	 & 27.3 & 	48.3 &  $-$ \\
          AdvEnt \cite{vu2019advent} & 85.6 & 42.2 & 79.7 & 8.7 & 0.4 & 25.9 & 5.4 & 8.1 & 80.4 & 84.1 & 57.9 & 23.8 & 73.3 & 36.4 & 14.2 & 33.0 & 48.0 & 41.2 \\
		ASA~\cite{zhou2020affinity} & \textbf{91.2} & 48.5 & 80.4 & 3.7 & 0.3 & 21.7 & 5.5 & 5.2 & 79.5 & 83.6 & 56.4 & 21.0 & 80.3 & 36.2 & 20.0 & 32.9 & 49.3 & 41.7 \\
		CBST \cite{YangZou2018UnsupervisedDA} & 68.0 & 29.9 & 76.3 & 10.8 & 1.4 & 33.9 & 22.8 & 29.5 & 77.6 & 78.3 & 60.6 & 28.3 & 81.6 & 23.5 & 18.8 & 39.8 & 48.9 & 42.6 \\
		MRNet~\cite{zheng2019unsupervised} & 82.0 & 36.5 & 80.4 & 4.2 & 0.4 & 33.7 & 18.0 & 13.4 & 81.1 & 80.8 & 61.3 & 21.7 & 84.4 & 32.4 & 14.8 & 45.7 & 50.2 & 43.2 \\
		MRKLD \cite{zou2019confidence} & 67.7 & 32.2 & 73.9 & 10.7 & 1.6 & 37.4 & 22.2 & 31.2 & 80.8 & 80.5 & 60.8 & 29.1 & 82.8 & 25.0 & 19.4 & 45.3 & 50.1 & 43.8 \\
		CCM~\cite{li2020content} & 79.6 & 36.4 & 80.6 & 13.3 & 0.3 & 25.5 & 22.4 & 14.9 & 81.8 & 77.4 & 56.8 & 25.9 & 80.7 & 45.3 & 29.9 &  52.0 & 52.9 & 45.2 \\
		Uncertainty~\cite{zheng2021rectifying} & 87.6 & 41.9 & 83.1 & 14.7 & 1.7 & 36.2 & 31.3 & 19.9 & 81.6 & 80.6 & 63.0 & 21.8 & 86.2 & 40.7 & 23.6 & 53.1 & 54.9 & 47.9 \\
		BL~\cite{li2019bidirectional} & 86.0 & 46.7 & 80.3 & $-$ & $-$ & $-$ & 14.1 & 11.6 & 79.2 & 81.3 & 54.1 & 27.9 & 73.7 & 42.2 & 25.7 & 45.3 & 51.4 & $-$ \\
		DT~\cite{wang2020differential} & 83.0 & 44.0 & 80.3 & $-$ & $-$ & $-$ & 17.1 & 15.8 & 80.5 & 81.8 & 59.9 & 33.1 & 70.2 & 37.3 & 28.5 & 45.8 & 52.1 &  $-$ \\
		IAST~\cite{KeMei2020InstanceAS} & 81.9 & 41.5 & 83.3 & 17.7 & 4.6 & 32.3 & 30.9 & 28.8 & 83.4 & 85.0 & 65.5 & 30.8 & 86.5 & 38.2 & 33.1 & 52.7 & 49.8 & - \\
		DAFormer \cite{hoyer2022daformer} & 84.5 & 40.7 & 88.4 & 41.5 & 6.5 & 50.0 & 55.0 & 54.6 & 86.0 & 89.8 & 73.2 & 48.2 & 87.2 & 53.2 & 53.9 & 61.7 & 67.4 & 60.9 \\
        CAMix \cite{zhou2022context} & 87.4 & 47.5 & 88.8 & $-$ & $-$ & $-$ & 55.2 & 55.4 & 87.0 & 91.7 & 72.0 & 49.3 & 86.9 & 57.0 & 57.5 & 63.6 & 69.2 & $-$ \\
        HRDA \cite{hoyer2022hrda} & 85.2 & 47.7 & 88.8 & 49.5 & 4.8 & 57.2 & 65.7 & 60.9 & 85.3 & 92.9 & 79.4 & 52.8 & 89.0 & 64.7 & 63.9 & 64.9 & 72.4 & 65.8 \\
        MIC \cite{hoyer2022mic} & 86.6 & 50.5 & 89.3 & 47.9 & 7.8 & 59.4 & 66.7 & 63.4 & 87.1 & \textbf{94.6} & \textbf{81.0} & \textbf{58.9
        }& 90.1 & 61.9 & \textbf{67.1} & 64.3 & 74.0 & 67.3 \\
        \hline
		DADA~\cite{vu2019dada} & 89.2 & 44.8 & 81.4 & 6.8 & 0.3 & 26.2 & 8.6 & 11.1 & 81.8 & 84.0 & 54.7 & 19.3 & 79.7 & 40.7 & 14.0 & 38.8 & 49.8 & 42.6 \\
        CorDA$^\dagger$~\cite{wang2021domain} & 93.3 & 61.6 & 85.3 & 19.6 & 5.1 & 37.8 & 36.6 & 42.8 & 84.9 & 90.4 & 69.7 & 41.8 & 85.6 & 38.4 & 32.6 & 53.9 & 62.8 & 55.0  \\
        Ours$^\dagger$ & \textbf{93.4} & \textbf{63.1} & \textbf{89.8} & \textbf{51.1} & \textbf{9.1} & \textbf{61.4} & \textbf{66.9} & \textbf{64.0} & \textbf{88.0} & 94.5 & 80.9 & 56.6 & \textbf{90.9} & \textbf{68.5} & 63.7 & \textbf{66.6} & \textbf{75.9} & \textbf{69.3} \\
		\shline
	\end{tabular}
	}
      \noindent\raggedright\footnotesize{$^\dagger$: Training with depth data.}
\end{table*}

\begin{table}[t]
    \centering
    \caption{Ablation study of different components of our proposed framework on GTA$\rightarrow$Cityscapes. The results are averaged over 3 random seeds.}
    \label{table:ablation}
    \small
    \resizebox{\linewidth}{!}{
    \begin{tabular}{cccccccc}
    \shline
    Method & ST Base. & Naive Mix. & DCF. & AFO. (CA)& AFO. (Trans + MMC)& mIoU$\uparrow$ \\
    \hline 
    M1 & $\checkmark$ &  & &  & & $73.1$  \\
    M2 & $\checkmark$ & $\checkmark$ &  & & & $76.0$  \\
    M3 & $\checkmark$ & $\checkmark$ & $\checkmark$ & &   & $77.1$  \\
    M4 & $\checkmark$ & $\checkmark$ & $\checkmark$ &  $\checkmark$&  &  $77.3$  \\
    M5 & $\checkmark$ & $\checkmark$ & $\checkmark$ &  & $\checkmark$ &  $77.7$  \\
    \shline
    \end{tabular}}
\end{table}

\vspace{-0in}
\begin{table}[]
     \small
    \centering
    \caption{Compatibility of the proposed method on different UDA methods and backbones on GTA$\rightarrow$Cityscapes. Our results are averaged over 3 random seeds.}
    \vspace{-0in}
    \begin{tabular}{llccc}
         \shline Backbone & UDA Method & w/o  & w/  & Diff. \\
         \hline 
         DeepLabV2 \cite{chen2017deeplab} & DACS \cite{WilhelmTranheden2020DACSDA} & $52.1$ & $56.2$ & $+4.1$ \\
         DAFormer \cite{hoyer2022daformer} & DAFormer \cite{hoyer2022daformer} & $68.3$ & $71.5$ & $+3.2$\\
         DAFormer \cite{hoyer2022daformer} & HRDA \cite{hoyer2022hrda} & $73.8$ & $76.1$ & $+2.3$ \\
         DAFormer \cite{hoyer2022daformer} & MIC \cite{hoyer2022mic} & $75.9$ & $77.7$ & $+1.8$ \\
         \shline
    \end{tabular}
    \vspace{-0in}
    \label{tab4}
\end{table}

\noindent\textbf{Results on Synthia$\rightarrow$Cityscapes.}
We show our results on SYNTHIA $\rightarrow$ Cityscapes in Table \ref{table:gtacity} and the results show the consistent performance improvement of our method, increasing from 67.3 to 69.3 (+2.0 mIoU) compared to the state-of-the-art method MIC \cite{hoyer2022mic}. Especially, our method significantly increases the IoU performance of the challenging class ``SideWalk'' from 50.5 to 63.1 (+12.6 mIoU). It is also noticeable that our method remains competitive in segmenting most individual classes and yields a significant increase of +6.8 on ``Road'', +6.6 on ``Bus'', +3.9 on ``Pole'', +3.7 on ``Road'', +3.2 on ``Wall'' and +2.9 on ``Truck''.

\subsection{Ablation Study and Further Disccussion}
\noindent\textbf{Ablation Study on Different Scene Adaptation Frameworks.} We combine our method with different scene adaptation architectures on GTA$\rightarrow$Cityscapes. Table \ref{tab4} shows that our method achieves consistent and significant improvements across different methods with different network architectures. Firstly, our method improves the state-of-the-art performance by +1.8 mIoU. Then we evaluate the proposed method on two strong methods based on transformer backbone, yielding +3.2 mIoU and +2.3 mIoU performance increase on DAFormer~\cite{hoyer2022daformer} and HRDA~\cite{hoyer2022hrda}, respectively. Secondly, we evaluate our method on DeepLabV2~\cite{chen2017deeplab} architecture with ResNet-101~\cite{he2016deep} backbone. We show that we improve the performance of the CNN-based cross-domain mixing method, \ie, DACS by +4.1 mIoU. The ablation study verifies the effectiveness of our method in leveraging depth information to enhance cross-domain mixing not only on Transformer-based networks but also on CNN-based architecture.

\noindent\textbf{Ablation Study on Different Components of the Proposed Method.} In order to verify the effectiveness of our proposed components, we train four different models from M1 to M4 and show the result in Table \ref{table:ablation}. ``ST Base'' means the self training baseline with semantic segmentation branch and depth regression branch. ``Naive Mix" denotes the cross-domain mixing strategy. ``DCF'' represents the proposed depth-aware mixing (Depth-guided Contextual Filter). ``AFO" denotes the proposed Adaptive Feature Optimization module and we used two different method to perform AFO. Firstly, we leverage channel attention (CA) that could select useful information along the channel dimension to perform the feature optimization. In this method, the fused feature is adaptively optimized by SENet \cite{hu2018squeeze}, the output is a weighted vector which is multiplied back to the visual and depth feature. We leavrage ``AFO (CA)'' to denote this method. Secondly, we leverage the iterative use of transformer block to adaptively optimize the multi-task feature. In this case, the output of the transformer block is a weighted map. The Multimodal Communication (MMC) module is then used to incorporate rich knowledge from the depth prediction. We denote this method as ``AFO (Trans + MMC)''.
M1 is the self-training baseline with depth regression based on DAFormer architecture. M2 adds the cross-domain mixing strategy for improvement and shows a competitive result of 76.0 mIoU. M3 is the model with the Depth-guided Contextual Filter, increasing the performance from 76.0 to 77.1 mIoU (+1.1 mIoU), which demonstrates the effectiveness of transferring the mixed training images to real-world layout with the help of the depth information. M4 adds the multi-task framework that leverages Channel Attention (CA) mechanism to fuse the discriminative depth feature into the visual feature. The segmentation result is increased by a small margin (+0.2 mIoU), which means CA could help the network to adaptively learn to focus or to ignore information from the auxiliary task to some extent. M5 is our proposed depth-aware multi-task model with both Depth-guided Contextual Filter and Adaptive Feature Optimization (AFO) module. Compared to M3, M5 has a mIoU increase of +0.6 from 77.1 to 77.7, which shows the effectiveness of multi-modal feature optimization using transformers to facilitate contextual learning.

\noindent\textbf{Ablation study on GTA+SYNTHIA $\rightarrow$ Cityscapes.} 
We evaluate the proposed method on multi-source domains setting and report the quantitative result on GTA+SYNTHIA $\rightarrow$ Cityscapes. With multi-source domain data, the model can be trained more robust to the unlabelled target environment. We adopt DACS~\cite{WilhelmTranheden2020DACSDA} as our baseline with 52.1 mIoU (Only GTA) performance shown in Table~\ref{table:cnn}. With more source-domain data, the model yields a better result of 54.2 mIoU. Then, we can observe that our method yields a larger improvement from 54.2 to 56.7 mIoU, demonstrating that the proposed model could adapt multi-domain depth to the target domain and hence increase performance.

\begin{table}[t]
    \centering
    \caption{Quantitative results on GTA+SYNTHIA → Cityscapes. The performance is provided as mIoU in $\%$.}
    \label{table:cnn}
    \footnotesize
    \begin{tabular}{c|c|c}
    \shline 
    Method & mIoU (\%) & $\Delta$ mIoU (\%) \\
    \hline
    Baseline (Single Source) & 52.1 & - \\
    Multi Source & 54.2 & +2.1 \\
    Adaboost~\cite{zheng2022adaptive} & 50.8 & - \\
    Multi Source + Depth & 56.7 & +4.6 \\
    \shline
    \end{tabular}
    \vspace{-0in}
\end{table}

\section{Conclusion}
In this work, we introduce a new depth-aware scene adaptation framework that effectively leverages the guidance of depth to enhance data augmentation and contextual learning. The proposed framework not only explicitly refines the cross-domain mixing by stimulating real-world layouts with the guidance of depth distributions of objects, but also introduced a cross-task encoder that adaptively optimizes the multi-task feature and focused on the discriminative depth feature to help contextual learning. By integrating our depth-aware framework into existing self-training methods based on either transformer or CNN, we achieve state-of-the-art performance on two widely used benchmarks and a significant improvement on small-scale categories. Extensive experimental results verify our motivation to transfer the training images to real-world layouts and demonstrate the effectiveness of our multi-task framework in improving scene adaptation performance.

\section*{Acknowledgement}
The paper is supported by Start-up Research Grant at the University of Macau (SRG2024-00002-FST), and Institute of Collaborative Innovation, University of Macau.

{\small
\bibliographystyle{ACM-Reference-Format}
\bibliography{egbib}


\begin{thebibliography}{104}


\ifx \showCODEN    \undefined \def \showCODEN     #1{\unskip}     \fi
\ifx \showDOI      \undefined \def \showDOI       #1{#1}\fi
\ifx \showISBNx    \undefined \def \showISBNx     #1{\unskip}     \fi
\ifx \showISBNxiii \undefined \def \showISBNxiii  #1{\unskip}     \fi
\ifx \showISSN     \undefined \def \showISSN      #1{\unskip}     \fi
\ifx \showLCCN     \undefined \def \showLCCN      #1{\unskip}     \fi
\ifx \shownote     \undefined \def \shownote      #1{#1}          \fi
\ifx \showarticletitle \undefined \def \showarticletitle #1{#1}   \fi
\ifx \showURL      \undefined \def \showURL       {\relax}        \fi
\providecommand\bibfield[2]{#2}
\providecommand\bibinfo[2]{#2}
\providecommand\natexlab[1]{#1}
\providecommand\showeprint[2][]{arXiv:#2}

\bibitem[Araslanov and Roth(2021)]%
        {araslanov2021self}
\bibfield{author}{\bibinfo{person}{Nikita Araslanov} {and} \bibinfo{person}{Stefan Roth}.} \bibinfo{year}{2021}\natexlab{}.
\newblock \showarticletitle{Self-supervised augmentation consistency for adapting semantic segmentation}. In \bibinfo{booktitle}{\emph{CVPR}}.
\newblock


\bibitem[Badrinarayanan et~al\mbox{.}(2017)]%
        {badrinarayanan2017segnet}
\bibfield{author}{\bibinfo{person}{Vijay Badrinarayanan}, \bibinfo{person}{Alex Kendall}, {and} \bibinfo{person}{Roberto Cipolla}.} \bibinfo{year}{2017}\natexlab{}.
\newblock \showarticletitle{Segnet: A deep convolutional encoder-decoder architecture for image segmentation}.
\newblock \bibinfo{journal}{\emph{IEEE transactions on pattern analysis and machine intelligence}} \bibinfo{volume}{39}, \bibinfo{number}{12} (\bibinfo{year}{2017}), \bibinfo{pages}{2481--2495}.
\newblock


\bibitem[Cardace et~al\mbox{.}(2022)]%
        {cardace2022plugging}
\bibfield{author}{\bibinfo{person}{Adriano Cardace}, \bibinfo{person}{Luca De~Luigi}, \bibinfo{person}{Pierluigi~Zama Ramirez}, \bibinfo{person}{Samuele Salti}, {and} \bibinfo{person}{Luigi Di~Stefano}.} \bibinfo{year}{2022}\natexlab{}.
\newblock \showarticletitle{Plugging self-supervised monocular depth into unsupervised domain adaptation for semantic segmentation}. In \bibinfo{booktitle}{\emph{CVPR}}.
\newblock


\bibitem[Chen et~al\mbox{.}(2022a)]%
        {chen2022reusing}
\bibfield{author}{\bibinfo{person}{Lin Chen}, \bibinfo{person}{Huaian Chen}, \bibinfo{person}{Zhixiang Wei}, \bibinfo{person}{Xin Jin}, \bibinfo{person}{Xiao Tan}, \bibinfo{person}{Yi Jin}, {and} \bibinfo{person}{Enhong Chen}.} \bibinfo{year}{2022}\natexlab{a}.
\newblock \showarticletitle{Reusing the task-specific classifier as a discriminator: Discriminator-free adversarial domain adaptation}. In \bibinfo{booktitle}{\emph{CVPR}}.
\newblock


\bibitem[Chen et~al\mbox{.}(2022b)]%
        {chen2022deliberated}
\bibfield{author}{\bibinfo{person}{Lin Chen}, \bibinfo{person}{Zhixiang Wei}, \bibinfo{person}{Xin Jin}, \bibinfo{person}{Huaian Chen}, \bibinfo{person}{Miao Zheng}, \bibinfo{person}{Kai Chen}, {and} \bibinfo{person}{Yi Jin}.} \bibinfo{year}{2022}\natexlab{b}.
\newblock \showarticletitle{Deliberated Domain Bridging for Domain Adaptive Semantic Segmentation}. In \bibinfo{booktitle}{\emph{NeurIPS}}.
\newblock


\bibitem[Chen et~al\mbox{.}(2017)]%
        {chen2017deeplab}
\bibfield{author}{\bibinfo{person}{Liang-Chieh Chen}, \bibinfo{person}{George Papandreou}, \bibinfo{person}{Iasonas Kokkinos}, \bibinfo{person}{Kevin Murphy}, {and} \bibinfo{person}{Alan~L Yuille}.} \bibinfo{year}{2017}\natexlab{}.
\newblock \showarticletitle{Deeplab: Semantic image segmentation with deep convolutional nets, atrous convolution, and fully connected crfs}.
\newblock \bibinfo{journal}{\emph{IEEE transactions on pattern analysis and machine intelligence}} \bibinfo{volume}{40}, \bibinfo{number}{4} (\bibinfo{year}{2017}), \bibinfo{pages}{834--848}.
\newblock


\bibitem[Chen et~al\mbox{.}(2024a)]%
        {chen2024uahoi}
\bibfield{author}{\bibinfo{person}{Mu Chen}, \bibinfo{person}{Minghan Chen}, {and} \bibinfo{person}{Yi Yang}.} \bibinfo{year}{2024}\natexlab{a}.
\newblock \showarticletitle{UAHOI: Uncertainty-aware robust interaction learning for HOI detection}.
\newblock \bibinfo{journal}{\emph{Computer Vision and Image Understanding}} (\bibinfo{year}{2024}), \bibinfo{pages}{104091}.
\newblock


\bibitem[Chen et~al\mbox{.}(2024b)]%
        {chen2024general}
\bibfield{author}{\bibinfo{person}{Mu Chen}, \bibinfo{person}{Liulei Li}, \bibinfo{person}{Wenguan Wang}, \bibinfo{person}{Ruijie Quan}, {and} \bibinfo{person}{Yi Yang}.} \bibinfo{year}{2024}\natexlab{b}.
\newblock \showarticletitle{General and Task-Oriented Video Segmentation}. In \bibinfo{booktitle}{\emph{ECCV}}.
\newblock


\bibitem[Chen et~al\mbox{.}(2019b)]%
        {chen2019domain}
\bibfield{author}{\bibinfo{person}{Minghao Chen}, \bibinfo{person}{Hongyang Xue}, {and} \bibinfo{person}{Deng Cai}.} \bibinfo{year}{2019}\natexlab{b}.
\newblock \showarticletitle{Domain adaptation for semantic segmentation with maximum squares loss}. In \bibinfo{booktitle}{\emph{ICCV}}.
\newblock


\bibitem[Chen et~al\mbox{.}(2023)]%
        {chen2023pipa}
\bibfield{author}{\bibinfo{person}{Mu Chen}, \bibinfo{person}{Zhedong Zheng}, \bibinfo{person}{Yi Yang}, {and} \bibinfo{person}{Tat-Seng Chua}.} \bibinfo{year}{2023}\natexlab{}.
\newblock \showarticletitle{Pipa: Pixel-and patch-wise self-supervised learning for domain adaptative semantic segmentation}. In \bibinfo{booktitle}{\emph{ACM MM}}. \bibinfo{pages}{1905--1914}.
\newblock


\bibitem[Chen et~al\mbox{.}(2019a)]%
        {chen2019learning}
\bibfield{author}{\bibinfo{person}{Yuhua Chen}, \bibinfo{person}{Wen Li}, \bibinfo{person}{Xiaoran Chen}, {and} \bibinfo{person}{Luc~Van Gool}.} \bibinfo{year}{2019}\natexlab{a}.
\newblock \showarticletitle{Learning semantic segmentation from synthetic data: A geometrically guided input-output adaptation approach}. In \bibinfo{booktitle}{\emph{CVPR}}.
\newblock


\bibitem[Cheng et~al\mbox{.}(2023)]%
        {cheng2023segment}
\bibfield{author}{\bibinfo{person}{Yangming Cheng}, \bibinfo{person}{Liulei Li}, \bibinfo{person}{Yuanyou Xu}, \bibinfo{person}{Xiaodi Li}, \bibinfo{person}{Zongxin Yang}, \bibinfo{person}{Wenguan Wang}, {and} \bibinfo{person}{Yi Yang}.} \bibinfo{year}{2023}\natexlab{}.
\newblock \showarticletitle{Segment and track anything}.
\newblock \bibinfo{journal}{\emph{arXiv preprint arXiv:2305.06558}} (\bibinfo{year}{2023}).
\newblock


\bibitem[Choi et~al\mbox{.}(2019)]%
        {choi2019self}
\bibfield{author}{\bibinfo{person}{Jaehoon Choi}, \bibinfo{person}{Taekyung Kim}, {and} \bibinfo{person}{Changick Kim}.} \bibinfo{year}{2019}\natexlab{}.
\newblock \showarticletitle{Self-ensembling with gan-based data augmentation for domain adaptation in semantic segmentation}. In \bibinfo{booktitle}{\emph{ICCV}}.
\newblock


\bibitem[Cordts et~al\mbox{.}(2016)]%
        {MariusCordts2016TheCD}
\bibfield{author}{\bibinfo{person}{Marius Cordts}, \bibinfo{person}{Mohamed Omran}, \bibinfo{person}{Sebastian Ramos}, \bibinfo{person}{Timo Rehfeld}, \bibinfo{person}{Markus Enzweiler}, \bibinfo{person}{Rodrigo Benenson}, \bibinfo{person}{Uwe Franke}, \bibinfo{person}{Stefan Roth}, {and} \bibinfo{person}{Bernt Schiele}.} \bibinfo{year}{2016}\natexlab{}.
\newblock \showarticletitle{The cityscapes dataset for semantic urban scene understanding}. In \bibinfo{booktitle}{\emph{CVPR}}.
\newblock


\bibitem[Ding et~al\mbox{.}(2024)]%
        {ding2024clustering}
\bibfield{author}{\bibinfo{person}{Yuhang Ding}, \bibinfo{person}{Liulei Li}, \bibinfo{person}{Wenguan Wang}, {and} \bibinfo{person}{Yi Yang}.} \bibinfo{year}{2024}\natexlab{}.
\newblock \showarticletitle{Clustering propagation for universal medical image segmentation}. In \bibinfo{booktitle}{\emph{CVPR}}.
\newblock


\bibitem[Feng et~al\mbox{.}(2021)]%
        {feng2021complementary}
\bibfield{author}{\bibinfo{person}{Hao Feng}, \bibinfo{person}{Minghao Chen}, \bibinfo{person}{Jinming Hu}, \bibinfo{person}{Dong Shen}, \bibinfo{person}{Haifeng Liu}, {and} \bibinfo{person}{Deng Cai}.} \bibinfo{year}{2021}\natexlab{}.
\newblock \showarticletitle{Complementary Pseudo Labels for Unsupervised Domain Adaptation On Person Re-Identification}.
\newblock \bibinfo{journal}{\emph{IEEE Transactions on Image Processing}}  \bibinfo{volume}{30} (\bibinfo{year}{2021}), \bibinfo{pages}{2898--2907}.
\newblock


\bibitem[Gholami et~al\mbox{.}(2020)]%
        {gholami2020unsupervised}
\bibfield{author}{\bibinfo{person}{Behnam Gholami}, \bibinfo{person}{Pritish Sahu}, \bibinfo{person}{Ognjen Rudovic}, \bibinfo{person}{Konstantinos Bousmalis}, {and} \bibinfo{person}{Vladimir Pavlovic}.} \bibinfo{year}{2020}\natexlab{}.
\newblock \showarticletitle{Unsupervised multi-target domain adaptation: An information theoretic approach}.
\newblock \bibinfo{journal}{\emph{IEEE Transactions on Image Processing}}  \bibinfo{volume}{29} (\bibinfo{year}{2020}), \bibinfo{pages}{3993--4002}.
\newblock


\bibitem[Godard et~al\mbox{.}(2019)]%
        {godard2019digging}
\bibfield{author}{\bibinfo{person}{Cl{\'e}ment Godard}, \bibinfo{person}{Oisin Mac~Aodha}, \bibinfo{person}{Michael Firman}, {and} \bibinfo{person}{Gabriel~J Brostow}.} \bibinfo{year}{2019}\natexlab{}.
\newblock \showarticletitle{Digging into self-supervised monocular depth estimation}. In \bibinfo{booktitle}{\emph{ICCV}}.
\newblock


\bibitem[Guo et~al\mbox{.}(2021)]%
        {ShaohuaGuo2021LabelFreeRC}
\bibfield{author}{\bibinfo{person}{Shaohua Guo}, \bibinfo{person}{Qianyu Zhou}, \bibinfo{person}{Ye Zhou}, \bibinfo{person}{Qiqi Gu}, \bibinfo{person}{Junshu Tang}, \bibinfo{person}{Zhengyang Feng}, {and} \bibinfo{person}{Lizhuang Ma}.} \bibinfo{year}{2021}\natexlab{}.
\newblock \showarticletitle{Label-free regional consistency for image-to-image translation}. In \bibinfo{booktitle}{\emph{ICME}}.
\newblock


\bibitem[He et~al\mbox{.}(2016)]%
        {he2016deep}
\bibfield{author}{\bibinfo{person}{Kaiming He}, \bibinfo{person}{Xiangyu Zhang}, \bibinfo{person}{Shaoqing Ren}, {and} \bibinfo{person}{Jian Sun}.} \bibinfo{year}{2016}\natexlab{}.
\newblock \showarticletitle{Deep residual learning for image recognition}. In \bibinfo{booktitle}{\emph{CVPR}}.
\newblock


\bibitem[Hoffman et~al\mbox{.}(2018)]%
        {hoffman2018cycada}
\bibfield{author}{\bibinfo{person}{Judy Hoffman}, \bibinfo{person}{Eric Tzeng}, \bibinfo{person}{Taesung Park}, \bibinfo{person}{Jun-Yan Zhu}, \bibinfo{person}{Phillip Isola}, \bibinfo{person}{Kate Saenko}, \bibinfo{person}{Alexei Efros}, {and} \bibinfo{person}{Trevor Darrell}.} \bibinfo{year}{2018}\natexlab{}.
\newblock \showarticletitle{Cycada: Cycle-consistent adversarial domain adaptation}. In \bibinfo{booktitle}{\emph{ICML}}.
\newblock


\bibitem[Hoyer et~al\mbox{.}(2022a)]%
        {hoyer2022daformer}
\bibfield{author}{\bibinfo{person}{Lukas Hoyer}, \bibinfo{person}{Dengxin Dai}, {and} \bibinfo{person}{Luc Van~Gool}.} \bibinfo{year}{2022}\natexlab{a}.
\newblock \showarticletitle{Daformer: Improving network architectures and training strategies for domain-adaptive semantic segmentation}. In \bibinfo{booktitle}{\emph{CVPR}}.
\newblock


\bibitem[Hoyer et~al\mbox{.}(2022b)]%
        {hoyer2022hrda}
\bibfield{author}{\bibinfo{person}{Lukas Hoyer}, \bibinfo{person}{Dengxin Dai}, {and} \bibinfo{person}{Luc Van~Gool}.} \bibinfo{year}{2022}\natexlab{b}.
\newblock \showarticletitle{HRDA: Context-Aware High-Resolution Domain-Adaptive Semantic Segmentation}. In \bibinfo{booktitle}{\emph{ECCV}}.
\newblock


\bibitem[Hoyer et~al\mbox{.}(2023)]%
        {hoyer2022mic}
\bibfield{author}{\bibinfo{person}{Lukas Hoyer}, \bibinfo{person}{Dengxin Dai}, \bibinfo{person}{Haoran Wang}, {and} \bibinfo{person}{Luc Van~Gool}.} \bibinfo{year}{2023}\natexlab{}.
\newblock \showarticletitle{{MIC}: Masked Image Consistency for Context-Enhanced Domain Adaptation}. In \bibinfo{booktitle}{\emph{CVPR}}.
\newblock


\bibitem[Hu et~al\mbox{.}(2018)]%
        {hu2018squeeze}
\bibfield{author}{\bibinfo{person}{Jie Hu}, \bibinfo{person}{Li Shen}, {and} \bibinfo{person}{Gang Sun}.} \bibinfo{year}{2018}\natexlab{}.
\newblock \showarticletitle{Squeeze-and-excitation networks}. In \bibinfo{booktitle}{\emph{CVPR}}.
\newblock


\bibitem[Huang et~al\mbox{.}(2022)]%
        {huang2022category}
\bibfield{author}{\bibinfo{person}{Jiaxing Huang}, \bibinfo{person}{Dayan Guan}, \bibinfo{person}{Aoran Xiao}, \bibinfo{person}{Shijian Lu}, {and} \bibinfo{person}{Ling Shao}.} \bibinfo{year}{2022}\natexlab{}.
\newblock \showarticletitle{Category contrast for unsupervised domain adaptation in visual tasks}. In \bibinfo{booktitle}{\emph{CVPR}}.
\newblock


\bibitem[Jiang et~al\mbox{.}(2022)]%
        {jiang2022prototypical}
\bibfield{author}{\bibinfo{person}{Zhengkai Jiang}, \bibinfo{person}{Yuxi Li}, \bibinfo{person}{Ceyuan Yang}, \bibinfo{person}{Peng Gao}, \bibinfo{person}{Yabiao Wang}, \bibinfo{person}{Ying Tai}, {and} \bibinfo{person}{Chengjie Wang}.} \bibinfo{year}{2022}\natexlab{}.
\newblock \showarticletitle{Prototypical Contrast Adaptation for Domain Adaptive Segmentation}. In \bibinfo{booktitle}{\emph{ECCV}}.
\newblock


\bibitem[Kanakis et~al\mbox{.}(2020)]%
        {kanakis2020reparameterizing}
\bibfield{author}{\bibinfo{person}{Menelaos Kanakis}, \bibinfo{person}{David Bruggemann}, \bibinfo{person}{Suman Saha}, \bibinfo{person}{Stamatios Georgoulis}, \bibinfo{person}{Anton Obukhov}, {and} \bibinfo{person}{Luc Van~Gool}.} \bibinfo{year}{2020}\natexlab{}.
\newblock \showarticletitle{Reparameterizing convolutions for incremental multi-task learning without task interference}. In \bibinfo{booktitle}{\emph{ECCV}}.
\newblock


\bibitem[Kim and Byun(2020)]%
        {kim2020learning}
\bibfield{author}{\bibinfo{person}{Myeongjin Kim} {and} \bibinfo{person}{Hyeran Byun}.} \bibinfo{year}{2020}\natexlab{}.
\newblock \showarticletitle{Learning texture invariant representation for domain adaptation of semantic segmentation}. In \bibinfo{booktitle}{\emph{CVPR}}.
\newblock


\bibitem[Laina et~al\mbox{.}(2016)]%
        {laina2016deeper}
\bibfield{author}{\bibinfo{person}{Iro Laina}, \bibinfo{person}{Christian Rupprecht}, \bibinfo{person}{Vasileios Belagiannis}, \bibinfo{person}{Federico Tombari}, {and} \bibinfo{person}{Nassir Navab}.} \bibinfo{year}{2016}\natexlab{}.
\newblock \showarticletitle{Deeper depth prediction with fully convolutional residual networks}. In \bibinfo{booktitle}{\emph{3DV}}.
\newblock


\bibitem[Lee et~al\mbox{.}(2018)]%
        {lee2018spigan}
\bibfield{author}{\bibinfo{person}{Kuan-Hui Lee}, \bibinfo{person}{German Ros}, \bibinfo{person}{Jie Li}, {and} \bibinfo{person}{Adrien Gaidon}.} \bibinfo{year}{2018}\natexlab{}.
\newblock \showarticletitle{Spigan: Privileged adversarial learning from simulation}.
\newblock \bibinfo{journal}{\emph{arXiv:1810.03756}} (\bibinfo{year}{2018}).
\newblock


\bibitem[Lee et~al\mbox{.}(2022)]%
        {lee2022adas}
\bibfield{author}{\bibinfo{person}{Seunghun Lee}, \bibinfo{person}{Wonhyeok Choi}, \bibinfo{person}{Changjae Kim}, \bibinfo{person}{Minwoo Choi}, {and} \bibinfo{person}{Sunghoon Im}.} \bibinfo{year}{2022}\natexlab{}.
\newblock \showarticletitle{ADAS: A Direct Adaptation Strategy for Multi-Target Domain Adaptive Semantic Segmentation}. In \bibinfo{booktitle}{\emph{CVPR}}.
\newblock


\bibitem[Li et~al\mbox{.}(2020)]%
        {li2020content}
\bibfield{author}{\bibinfo{person}{Guangrui Li}, \bibinfo{person}{Guoliang Kang}, \bibinfo{person}{Wu Liu}, \bibinfo{person}{Yunchao Wei}, {and} \bibinfo{person}{Yi Yang}.} \bibinfo{year}{2020}\natexlab{}.
\newblock \showarticletitle{Content-Consistent Matching for Domain Adaptive Semantic Segmentation}. In \bibinfo{booktitle}{\emph{ECCV}}.
\newblock


\bibitem[Li et~al\mbox{.}(2023a)]%
        {li2023logicseg}
\bibfield{author}{\bibinfo{person}{Liulei Li}, \bibinfo{person}{Wenguan Wang}, {and} \bibinfo{person}{Yi Yang}.} \bibinfo{year}{2023}\natexlab{a}.
\newblock \showarticletitle{Logicseg: Parsing visual semantics with neural logic learning and reasoning}. In \bibinfo{booktitle}{\emph{ICCV}}.
\newblock


\bibitem[Li et~al\mbox{.}(2023b)]%
        {li2023semantic}
\bibfield{author}{\bibinfo{person}{Liulei Li}, \bibinfo{person}{Wenguan Wang}, \bibinfo{person}{Tianfei Zhou}, \bibinfo{person}{Ruijie Quan}, {and} \bibinfo{person}{Yi Yang}.} \bibinfo{year}{2023}\natexlab{b}.
\newblock \showarticletitle{Semantic hierarchy-aware segmentation}.
\newblock \bibinfo{journal}{\emph{IEEE TPAMI}} (\bibinfo{year}{2023}).
\newblock


\bibitem[Li et~al\mbox{.}(2022b)]%
        {li2022deep}
\bibfield{author}{\bibinfo{person}{Liulei Li}, \bibinfo{person}{Tianfei Zhou}, \bibinfo{person}{Wenguan Wang}, \bibinfo{person}{Jianwu Li}, {and} \bibinfo{person}{Yi Yang}.} \bibinfo{year}{2022}\natexlab{b}.
\newblock \showarticletitle{Deep hierarchical semantic segmentation}. In \bibinfo{booktitle}{\emph{IEEE Conf. Comput. Vis. Pattern Recog.}}
\newblock


\bibitem[Li et~al\mbox{.}(2022a)]%
        {li2022class}
\bibfield{author}{\bibinfo{person}{Ruihuang Li}, \bibinfo{person}{Shuai Li}, \bibinfo{person}{Chenhang He}, \bibinfo{person}{Yabin Zhang}, \bibinfo{person}{Xu Jia}, {and} \bibinfo{person}{Lei Zhang}.} \bibinfo{year}{2022}\natexlab{a}.
\newblock \showarticletitle{Class-balanced pixel-level self-labeling for domain adaptive semantic segmentation}. In \bibinfo{booktitle}{\emph{CVPR}}.
\newblock


\bibitem[Li et~al\mbox{.}(2019)]%
        {li2019bidirectional}
\bibfield{author}{\bibinfo{person}{Yunsheng Li}, \bibinfo{person}{Lu Yuan}, {and} \bibinfo{person}{Nuno Vasconcelos}.} \bibinfo{year}{2019}\natexlab{}.
\newblock \showarticletitle{Bidirectional learning for domain adaptation of semantic segmentation}. In \bibinfo{booktitle}{\emph{CVPR}}.
\newblock


\bibitem[Lian et~al\mbox{.}(2019)]%
        {lian2019constructing}
\bibfield{author}{\bibinfo{person}{Qing Lian}, \bibinfo{person}{Fengmao Lv}, \bibinfo{person}{Lixin Duan}, {and} \bibinfo{person}{Boqing Gong}.} \bibinfo{year}{2019}\natexlab{}.
\newblock \showarticletitle{Constructing self-motivated pyramid curriculums for cross-domain semantic segmentation: A non-adversarial approach}. In \bibinfo{booktitle}{\emph{ICCV}}.
\newblock


\bibitem[Liang et~al\mbox{.}(2022)]%
        {liang2022gmmseg}
\bibfield{author}{\bibinfo{person}{Chen Liang}, \bibinfo{person}{Wenguan Wang}, \bibinfo{person}{Jiaxu Miao}, {and} \bibinfo{person}{Yi Yang}.} \bibinfo{year}{2022}\natexlab{}.
\newblock \showarticletitle{Gmmseg: Gaussian mixture based generative semantic segmentation models}. In \bibinfo{booktitle}{\emph{NeurIPS}}.
\newblock


\bibitem[Liang et~al\mbox{.}(2023)]%
        {liang2023clustseg}
\bibfield{author}{\bibinfo{person}{James~Chenhao Liang}, \bibinfo{person}{Tianfei Zhou}, \bibinfo{person}{Dongfang Liu}, {and} \bibinfo{person}{Wenguan Wang}.} \bibinfo{year}{2023}\natexlab{}.
\newblock \showarticletitle{CLUSTSEG: Clustering for Universal Segmentation}. In \bibinfo{booktitle}{\emph{ICML}}.
\newblock


\bibitem[Liu et~al\mbox{.}(2024)]%
        {liu2024high}
\bibfield{author}{\bibinfo{person}{Jinliang Liu}, \bibinfo{person}{Zhedong Zheng}, \bibinfo{person}{Zongxin Yang}, {and} \bibinfo{person}{Yi Yang}.} \bibinfo{year}{2024}\natexlab{}.
\newblock \showarticletitle{High Fidelity Makeup via 2D and 3D Identity Preservation Net}.
\newblock \bibinfo{journal}{\emph{ACM Transactions on Multimedia Computing, Communications and Applications}} (\bibinfo{year}{2024}).
\newblock


\bibitem[Liu et~al\mbox{.}(2021)]%
        {liu2021bapa}
\bibfield{author}{\bibinfo{person}{Yahao Liu}, \bibinfo{person}{Jinhong Deng}, \bibinfo{person}{Xinchen Gao}, \bibinfo{person}{Wen Li}, {and} \bibinfo{person}{Lixin Duan}.} \bibinfo{year}{2021}\natexlab{}.
\newblock \showarticletitle{Bapa-net: Boundary adaptation and prototype alignment for cross-domain semantic segmentation}. In \bibinfo{booktitle}{\emph{ICCV}}.
\newblock


\bibitem[Liu et~al\mbox{.}(2022)]%
        {liu2022undoing}
\bibfield{author}{\bibinfo{person}{Yahao Liu}, \bibinfo{person}{Jinhong Deng}, \bibinfo{person}{Jiale Tao}, \bibinfo{person}{Tong Chu}, \bibinfo{person}{Lixin Duan}, {and} \bibinfo{person}{Wen Li}.} \bibinfo{year}{2022}\natexlab{}.
\newblock \showarticletitle{Undoing the damage of label shift for cross-domain semantic segmentation}. In \bibinfo{booktitle}{\emph{CVPR}}.
\newblock


\bibitem[Long et~al\mbox{.}(2015)]%
        {long2015fully}
\bibfield{author}{\bibinfo{person}{Jonathan Long}, \bibinfo{person}{Evan Shelhamer}, {and} \bibinfo{person}{Trevor Darrell}.} \bibinfo{year}{2015}\natexlab{}.
\newblock \showarticletitle{Fully convolutional networks for semantic segmentation}. In \bibinfo{booktitle}{\emph{CVPR}}.
\newblock


\bibitem[Luo et~al\mbox{.}(2019a)]%
        {luo2019significance}
\bibfield{author}{\bibinfo{person}{Yawei Luo}, \bibinfo{person}{Ping Liu}, \bibinfo{person}{Tao Guan}, \bibinfo{person}{Junqing Yu}, {and} \bibinfo{person}{Yi Yang}.} \bibinfo{year}{2019}\natexlab{a}.
\newblock \showarticletitle{Significance-aware Information Bottleneck for Domain Adaptive Semantic Segmentation}. In \bibinfo{booktitle}{\emph{ICCV}}.
\newblock


\bibitem[Luo et~al\mbox{.}(2019b)]%
        {luo2019taking}
\bibfield{author}{\bibinfo{person}{Yawei Luo}, \bibinfo{person}{Liang Zheng}, \bibinfo{person}{Tao Guan}, \bibinfo{person}{Junqing Yu}, {and} \bibinfo{person}{Yi Yang}.} \bibinfo{year}{2019}\natexlab{b}.
\newblock \showarticletitle{Taking a closer look at domain shift: Category-level adversaries for semantics consistent domain adaptation}. In \bibinfo{booktitle}{\emph{CVPR}}.
\newblock


\bibitem[Mei et~al\mbox{.}(2020)]%
        {KeMei2020InstanceAS}
\bibfield{author}{\bibinfo{person}{Ke Mei}, \bibinfo{person}{Chuang Zhu}, \bibinfo{person}{Jiaqi Zou}, {and} \bibinfo{person}{Shanghang Zhang}.} \bibinfo{year}{2020}\natexlab{}.
\newblock \showarticletitle{Instance adaptive self-training for unsupervised domain adaptation}. In \bibinfo{booktitle}{\emph{ECCV}}.
\newblock


\bibitem[Melas-Kyriazi and Manrai(2021)]%
        {melas2021pixmatch}
\bibfield{author}{\bibinfo{person}{Luke Melas-Kyriazi} {and} \bibinfo{person}{Arjun~K Manrai}.} \bibinfo{year}{2021}\natexlab{}.
\newblock \showarticletitle{Pixmatch: Unsupervised domain adaptation via pixelwise consistency training}. In \bibinfo{booktitle}{\emph{CVPR}}.
\newblock


\bibitem[Olsson et~al\mbox{.}(2021)]%
        {olsson2021classmix}
\bibfield{author}{\bibinfo{person}{Viktor Olsson}, \bibinfo{person}{Wilhelm Tranheden}, \bibinfo{person}{Juliano Pinto}, {and} \bibinfo{person}{Lennart Svensson}.} \bibinfo{year}{2021}\natexlab{}.
\newblock \showarticletitle{Classmix: Segmentation-based data augmentation for semi-supervised learning}. In \bibinfo{booktitle}{\emph{WACV}}.
\newblock


\bibitem[Paszke et~al\mbox{.}(2017)]%
        {paszke2017automatic}
\bibfield{author}{\bibinfo{person}{Adam Paszke}, \bibinfo{person}{Sam Gross}, \bibinfo{person}{Soumith Chintala}, \bibinfo{person}{Gregory Chanan}, \bibinfo{person}{Edward Yang}, \bibinfo{person}{Zachary DeVito}, \bibinfo{person}{Zeming Lin}, \bibinfo{person}{Alban Desmaison}, \bibinfo{person}{Luca Antiga}, {and} \bibinfo{person}{Adam Lerer}.} \bibinfo{year}{2017}\natexlab{}.
\newblock \showarticletitle{Automatic differentiation in PyTorch}.
\newblock  (\bibinfo{year}{2017}).
\newblock


\bibitem[Rangwani et~al\mbox{.}(2022)]%
        {rangwani2022closer}
\bibfield{author}{\bibinfo{person}{Harsh Rangwani}, \bibinfo{person}{Sumukh~K Aithal}, \bibinfo{person}{Mayank Mishra}, \bibinfo{person}{Arihant Jain}, {and} \bibinfo{person}{Venkatesh~Babu Radhakrishnan}.} \bibinfo{year}{2022}\natexlab{}.
\newblock \showarticletitle{A closer look at smoothness in domain adversarial training}. In \bibinfo{booktitle}{\emph{ICML}}.
\newblock


\bibitem[Richter et~al\mbox{.}(2016)]%
        {StephanRRichter2016PlayingFD}
\bibfield{author}{\bibinfo{person}{Stephan~R Richter}, \bibinfo{person}{Vibhav Vineet}, \bibinfo{person}{Stefan Roth}, {and} \bibinfo{person}{Vladlen Koltun}.} \bibinfo{year}{2016}\natexlab{}.
\newblock \showarticletitle{Playing for data: Ground truth from computer games}. In \bibinfo{booktitle}{\emph{ECCV}}.
\newblock


\bibitem[Ros et~al\mbox{.}(2016)]%
        {ros2016synthia}
\bibfield{author}{\bibinfo{person}{German Ros}, \bibinfo{person}{Laura Sellart}, \bibinfo{person}{Joanna Materzynska}, \bibinfo{person}{David Vazquez}, {and} \bibinfo{person}{Antonio~M Lopez}.} \bibinfo{year}{2016}\natexlab{}.
\newblock \showarticletitle{The synthia dataset: A large collection of synthetic images for semantic segmentation of urban scenes}. In \bibinfo{booktitle}{\emph{CVPR}}.
\newblock


\bibitem[Saha et~al\mbox{.}(2021)]%
        {saha2021learning}
\bibfield{author}{\bibinfo{person}{Suman Saha}, \bibinfo{person}{Anton Obukhov}, \bibinfo{person}{Danda~Pani Paudel}, \bibinfo{person}{Menelaos Kanakis}, \bibinfo{person}{Yuhua Chen}, \bibinfo{person}{Stamatios Georgoulis}, {and} \bibinfo{person}{Luc Van~Gool}.} \bibinfo{year}{2021}\natexlab{}.
\newblock \showarticletitle{Learning to relate depth and semantics for unsupervised domain adaptation}. In \bibinfo{booktitle}{\emph{CVPR}}.
\newblock


\bibitem[Sakaridis et~al\mbox{.}(2018)]%
        {sakaridis2018model}
\bibfield{author}{\bibinfo{person}{Christos Sakaridis}, \bibinfo{person}{Dengxin Dai}, \bibinfo{person}{Simon Hecker}, {and} \bibinfo{person}{Luc Van~Gool}.} \bibinfo{year}{2018}\natexlab{}.
\newblock \showarticletitle{Model adaptation with synthetic and real data for semantic dense foggy scene understanding}. In \bibinfo{booktitle}{\emph{ECCV}}.
\newblock


\bibitem[Saporta et~al\mbox{.}(2021)]%
        {saporta2021multi}
\bibfield{author}{\bibinfo{person}{Antoine Saporta}, \bibinfo{person}{Tuan-Hung Vu}, \bibinfo{person}{Matthieu Cord}, {and} \bibinfo{person}{Patrick P{\'e}rez}.} \bibinfo{year}{2021}\natexlab{}.
\newblock \showarticletitle{Multi-target adversarial frameworks for domain adaptation in semantic segmentation}. In \bibinfo{booktitle}{\emph{ICCV}}.
\newblock


\bibitem[Shan et~al\mbox{.}(2020)]%
        {SHAN2020125}
\bibfield{author}{\bibinfo{person}{Yuhu Shan}, \bibinfo{person}{Chee~Meng Chew}, {and} \bibinfo{person}{Wen~Feng Lu}.} \bibinfo{year}{2020}\natexlab{}.
\newblock \showarticletitle{Semantic-aware short path adversarial training for cross-domain semantic segmentation}.
\newblock \bibinfo{journal}{\emph{Neurocomputing}}  \bibinfo{volume}{380} (\bibinfo{year}{2020}), \bibinfo{pages}{125--132}.
\newblock
\showISSN{0925-2312}
\urldef\tempurl%
\url{https://doi.org/10.1016/j.neucom.2019.11.008}
\showDOI{\tempurl}


\bibitem[Standley et~al\mbox{.}(2020)]%
        {standley2020tasks}
\bibfield{author}{\bibinfo{person}{Trevor Standley}, \bibinfo{person}{Amir Zamir}, \bibinfo{person}{Dawn Chen}, \bibinfo{person}{Leonidas Guibas}, \bibinfo{person}{Jitendra Malik}, {and} \bibinfo{person}{Silvio Savarese}.} \bibinfo{year}{2020}\natexlab{}.
\newblock \showarticletitle{Which tasks should be learned together in multi-task learning?}. In \bibinfo{booktitle}{\emph{ICML}}.
\newblock


\bibitem[Sun et~al\mbox{.}(2020)]%
        {sun2020mining}
\bibfield{author}{\bibinfo{person}{Guolei Sun}, \bibinfo{person}{Wenguan Wang}, \bibinfo{person}{Jifeng Dai}, {and} \bibinfo{person}{Luc Van~Gool}.} \bibinfo{year}{2020}\natexlab{}.
\newblock \showarticletitle{Mining cross-image semantics for weakly supervised semantic segmentation}. In \bibinfo{booktitle}{\emph{ECCV}}.
\newblock


\bibitem[Tranheden et~al\mbox{.}(2021)]%
        {WilhelmTranheden2020DACSDA}
\bibfield{author}{\bibinfo{person}{Wilhelm Tranheden}, \bibinfo{person}{Viktor Olsson}, \bibinfo{person}{Juliano Pinto}, {and} \bibinfo{person}{Lennart Svensson}.} \bibinfo{year}{2021}\natexlab{}.
\newblock \showarticletitle{Dacs: Domain adaptation via cross-domain mixed sampling}. In \bibinfo{booktitle}{\emph{WACV Workshop}}.
\newblock


\bibitem[Tsai et~al\mbox{.}(2018)]%
        {tsai2018learning}
\bibfield{author}{\bibinfo{person}{Yi-Hsuan Tsai}, \bibinfo{person}{Wei-Chih Hung}, \bibinfo{person}{Samuel Schulter}, \bibinfo{person}{Kihyuk Sohn}, \bibinfo{person}{Ming-Hsuan Yang}, {and} \bibinfo{person}{Manmohan Chandraker}.} \bibinfo{year}{2018}\natexlab{}.
\newblock \showarticletitle{Learning to adapt structured output space for semantic segmentation}. In \bibinfo{booktitle}{\emph{CVPR}}.
\newblock


\bibitem[Tsai et~al\mbox{.}(2019)]%
        {tsai2019domain}
\bibfield{author}{\bibinfo{person}{Yi-Hsuan Tsai}, \bibinfo{person}{Kihyuk Sohn}, \bibinfo{person}{Samuel Schulter}, {and} \bibinfo{person}{Manmohan Chandraker}.} \bibinfo{year}{2019}\natexlab{}.
\newblock \showarticletitle{Domain adaptation for structured output via discriminative patch representations}. In \bibinfo{booktitle}{\emph{ICCV}}.
\newblock


\bibitem[Vandenhende et~al\mbox{.}(2020)]%
        {vandenhende2020mti}
\bibfield{author}{\bibinfo{person}{Simon Vandenhende}, \bibinfo{person}{Stamatios Georgoulis}, {and} \bibinfo{person}{Luc Van~Gool}.} \bibinfo{year}{2020}\natexlab{}.
\newblock \showarticletitle{Mti-net: Multi-scale task interaction networks for multi-task learning}. In \bibinfo{booktitle}{\emph{ECCV}}.
\newblock


\bibitem[Vu et~al\mbox{.}(2019a)]%
        {vu2019advent}
\bibfield{author}{\bibinfo{person}{Tuan-Hung Vu}, \bibinfo{person}{Himalaya Jain}, \bibinfo{person}{Maxime Bucher}, \bibinfo{person}{Matthieu Cord}, {and} \bibinfo{person}{Patrick P{\'e}rez}.} \bibinfo{year}{2019}\natexlab{a}.
\newblock \showarticletitle{Advent: Adversarial entropy minimization for domain adaptation in semantic segmentation}. In \bibinfo{booktitle}{\emph{CVPR}}.
\newblock


\bibitem[Vu et~al\mbox{.}(2019b)]%
        {vu2019dada}
\bibfield{author}{\bibinfo{person}{Tuan-Hung Vu}, \bibinfo{person}{Himalaya Jain}, \bibinfo{person}{Maxime Bucher}, \bibinfo{person}{Matthieu Cord}, {and} \bibinfo{person}{Patrick P{\'e}rez}.} \bibinfo{year}{2019}\natexlab{b}.
\newblock \showarticletitle{Dada: Depth-aware domain adaptation in semantic segmentation}. In \bibinfo{booktitle}{\emph{ICCV}}.
\newblock


\bibitem[Wang et~al\mbox{.}(2024c)]%
        {DeBNet}
\bibfield{author}{\bibinfo{person}{Chao Wang}, \bibinfo{person}{Zhedong Zheng}, \bibinfo{person}{Ruijie Quan}, {and} \bibinfo{person}{Yi Yang}.} \bibinfo{year}{2024}\natexlab{c}.
\newblock \showarticletitle{Depth-aware blind image decomposition for real-world adverse weather recovery}. In \bibinfo{booktitle}{\emph{ACM MM}}.
\newblock


\bibitem[Wang et~al\mbox{.}(2020a)]%
        {HaoranWang2020ClassesMA}
\bibfield{author}{\bibinfo{person}{Haoran Wang}, \bibinfo{person}{Tong Shen}, \bibinfo{person}{Wei Zhang}, \bibinfo{person}{Ling-Yu Duan}, {and} \bibinfo{person}{Tao Mei}.} \bibinfo{year}{2020}\natexlab{a}.
\newblock \showarticletitle{Classes matter: A fine-grained adversarial approach to cross-domain semantic segmentation}. In \bibinfo{booktitle}{\emph{ECCV}}.
\newblock


\bibitem[Wang et~al\mbox{.}(2021a)]%
        {wang2021domain}
\bibfield{author}{\bibinfo{person}{Qin Wang}, \bibinfo{person}{Dengxin Dai}, \bibinfo{person}{Lukas Hoyer}, \bibinfo{person}{Luc Van~Gool}, {and} \bibinfo{person}{Olga Fink}.} \bibinfo{year}{2021}\natexlab{a}.
\newblock \showarticletitle{Domain adaptive semantic segmentation with self-supervised depth estimation}. In \bibinfo{booktitle}{\emph{ICCV}}.
\newblock


\bibitem[Wang et~al\mbox{.}(2024a)]%
        {wang2024disentangled}
\bibfield{author}{\bibinfo{person}{Shanshan Wang}, \bibinfo{person}{ALuSi}, \bibinfo{person}{Xun Yang}, \bibinfo{person}{Ke Xu}, \bibinfo{person}{Huibin Tan}, {and} \bibinfo{person}{Xingyi Zhang}.} \bibinfo{year}{2024}\natexlab{a}.
\newblock \showarticletitle{Dual-stream Feature Augmentation for Domain Generalization}. In \bibinfo{booktitle}{\emph{ACM MM}}.
\newblock


\bibitem[Wang et~al\mbox{.}(2023)]%
        {wang2023disentangled}
\bibfield{author}{\bibinfo{person}{Shanshan Wang}, \bibinfo{person}{Yiyang Chen}, \bibinfo{person}{Zhenwei He}, \bibinfo{person}{Xun Yang}, \bibinfo{person}{Mengzhu Wang}, \bibinfo{person}{Quanzeng You}, {and} \bibinfo{person}{Xingyi Zhang}.} \bibinfo{year}{2023}\natexlab{}.
\newblock \showarticletitle{Disentangled representation learning with causality for unsupervised domain adaptation}. In \bibinfo{booktitle}{\emph{ACM MM}}.
\newblock


\bibitem[Wang et~al\mbox{.}(2019)]%
        {wang2019zero}
\bibfield{author}{\bibinfo{person}{Wenguan Wang}, \bibinfo{person}{Xiankai Lu}, \bibinfo{person}{Jianbing Shen}, \bibinfo{person}{David~J Crandall}, {and} \bibinfo{person}{Ling Shao}.} \bibinfo{year}{2019}\natexlab{}.
\newblock \showarticletitle{Zero-shot video object segmentation via attentive graph neural networks}. In \bibinfo{booktitle}{\emph{ICCV}}.
\newblock


\bibitem[Wang et~al\mbox{.}(2024b)]%
        {wang2024visual}
\bibfield{author}{\bibinfo{person}{Wenguan Wang}, \bibinfo{person}{Yi Yang}, {and} \bibinfo{person}{Yunhe Pan}.} \bibinfo{year}{2024}\natexlab{b}.
\newblock \showarticletitle{Visual Knowledge in the Big Model Era: Retrospect and Prospect}.
\newblock \bibinfo{journal}{\emph{arXiv preprint arXiv:2404.04308}} (\bibinfo{year}{2024}).
\newblock


\bibitem[Wang et~al\mbox{.}(2021b)]%
        {WangWQZ021}
\bibfield{author}{\bibinfo{person}{Yaxiong Wang}, \bibinfo{person}{Yunchao Wei}, \bibinfo{person}{Xueming Qian}, \bibinfo{person}{Li Zhu}, {and} \bibinfo{person}{Yi Yang}.} \bibinfo{year}{2021}\natexlab{b}.
\newblock \showarticletitle{AINet: Association Implantation for Superpixel Segmentation}. In \bibinfo{booktitle}{\emph{2021 {IEEE/CVF} International Conference on Computer Vision, {ICCV} 2021, Montreal, QC, Canada, October 10-17, 2021}}. \bibinfo{publisher}{{IEEE}}, \bibinfo{pages}{7058--7067}.
\newblock
\urldef\tempurl%
\url{https://doi.org/10.1109/ICCV48922.2021.00699}
\showDOI{\tempurl}


\bibitem[Wang et~al\mbox{.}(2020b)]%
        {wang2020differential}
\bibfield{author}{\bibinfo{person}{Zhonghao Wang}, \bibinfo{person}{Mo Yu}, \bibinfo{person}{Yunchao Wei}, \bibinfo{person}{Rogerio Feris}, \bibinfo{person}{Jinjun Xiong}, \bibinfo{person}{Wen-mei Hwu}, \bibinfo{person}{Thomas~S Huang}, {and} \bibinfo{person}{Honghui Shi}.} \bibinfo{year}{2020}\natexlab{b}.
\newblock \showarticletitle{Differential treatment for stuff and things: A simple unsupervised domain adaptation method for semantic segmentation}. In \bibinfo{booktitle}{\emph{CVPR}}.
\newblock


\bibitem[Wu and Liu(2022)]%
        {wuunsupervised}
\bibfield{author}{\bibinfo{person}{Quanliang Wu} {and} \bibinfo{person}{Huajun Liu}.} \bibinfo{year}{2022}\natexlab{}.
\newblock \showarticletitle{Unsupervised Domain Adaptation for Semantic Segmentation using Depth Distribution}. In \bibinfo{booktitle}{\emph{NeurIPS}}.
\newblock


\bibitem[Wu et~al\mbox{.}(2019)]%
        {ZuxuanWu2019ACEAT}
\bibfield{author}{\bibinfo{person}{Zuxuan Wu}, \bibinfo{person}{Xin Wang}, \bibinfo{person}{Joseph~E Gonzalez}, \bibinfo{person}{Tom Goldstein}, {and} \bibinfo{person}{Larry~S Davis}.} \bibinfo{year}{2019}\natexlab{}.
\newblock \showarticletitle{Ace: Adapting to changing environments for semantic segmentation}. In \bibinfo{booktitle}{\emph{ICCV}}.
\newblock


\bibitem[Xie et~al\mbox{.}(2022)]%
        {BinhuiXie2022SePiCoSP}
\bibfield{author}{\bibinfo{person}{Binhui Xie}, \bibinfo{person}{Shuang Li}, \bibinfo{person}{Mingjia Li}, \bibinfo{person}{Chi~Harold Liu}, \bibinfo{person}{Gao Huang}, {and} \bibinfo{person}{Guoren Wang}.} \bibinfo{year}{2022}\natexlab{}.
\newblock \showarticletitle{SePiCo: Semantic-Guided Pixel Contrast for Domain Adaptive Semantic Segmentation}.
\newblock \bibinfo{journal}{\emph{arXiv:2204.08808}} (\bibinfo{year}{2022}).
\newblock


\bibitem[Xie et~al\mbox{.}(2021)]%
        {xie2021segformer}
\bibfield{author}{\bibinfo{person}{Enze Xie}, \bibinfo{person}{Wenhai Wang}, \bibinfo{person}{Zhiding Yu}, \bibinfo{person}{Anima Anandkumar}, \bibinfo{person}{Jose~M Alvarez}, {and} \bibinfo{person}{Ping Luo}.} \bibinfo{year}{2021}\natexlab{}.
\newblock \showarticletitle{SegFormer: Simple and efficient design for semantic segmentation with transformers}.
\newblock \bibinfo{journal}{\emph{NeurIPS}} (\bibinfo{year}{2021}).
\newblock


\bibitem[Xu et~al\mbox{.}(2018)]%
        {xu2018pad}
\bibfield{author}{\bibinfo{person}{Dan Xu}, \bibinfo{person}{Wanli Ouyang}, \bibinfo{person}{Xiaogang Wang}, {and} \bibinfo{person}{Nicu Sebe}.} \bibinfo{year}{2018}\natexlab{}.
\newblock \showarticletitle{Pad-net: Multi-tasks guided prediction-and-distillation network for simultaneous depth estimation and scene parsing}. In \bibinfo{booktitle}{\emph{CVPR}}.
\newblock


\bibitem[Yang et~al\mbox{.}(2020a)]%
        {JinyuYang2020LabelDrivenRF}
\bibfield{author}{\bibinfo{person}{Jinyu Yang}, \bibinfo{person}{Weizhi An}, \bibinfo{person}{Sheng Wang}, \bibinfo{person}{Xinliang Zhu}, \bibinfo{person}{Chaochao Yan}, {and} \bibinfo{person}{Junzhou Huang}.} \bibinfo{year}{2020}\natexlab{a}.
\newblock \showarticletitle{Label-driven reconstruction for domain adaptation in semantic segmentation}. In \bibinfo{booktitle}{\emph{ECCV}}.
\newblock


\bibitem[Yang et~al\mbox{.}(2020b)]%
        {yang2020adversarial}
\bibfield{author}{\bibinfo{person}{Jihan Yang}, \bibinfo{person}{Ruijia Xu}, \bibinfo{person}{Ruiyu Li}, \bibinfo{person}{Xiaojuan Qi}, \bibinfo{person}{Xiaoyong Shen}, \bibinfo{person}{Guanbin Li}, {and} \bibinfo{person}{Liang Lin}.} \bibinfo{year}{2020}\natexlab{b}.
\newblock \showarticletitle{An Adversarial Perturbation Oriented Domain Adaptation Approach for Semantic Segmentation}. In \bibinfo{booktitle}{\emph{AAAI}}.
\newblock


\bibitem[Yang et~al\mbox{.}(2024b)]%
        {yang2024dgl}
\bibfield{author}{\bibinfo{person}{Xiangpeng Yang}, \bibinfo{person}{Linchao Zhu}, \bibinfo{person}{Xiaohan Wang}, {and} \bibinfo{person}{Yi Yang}.} \bibinfo{year}{2024}\natexlab{b}.
\newblock \showarticletitle{DGL: Dynamic Global-Local Prompt Tuning for Text-Video Retrieval}. In \bibinfo{booktitle}{\emph{AAAI}}.
\newblock


\bibitem[Yang and Soatto(2020)]%
        {yang2020fda}
\bibfield{author}{\bibinfo{person}{Yanchao Yang} {and} \bibinfo{person}{Stefano Soatto}.} \bibinfo{year}{2020}\natexlab{}.
\newblock \showarticletitle{Fda: Fourier domain adaptation for semantic segmentation}. In \bibinfo{booktitle}{\emph{CVPR}}.
\newblock


\bibitem[Yang et~al\mbox{.}(2024a)]%
        {yang2024doraemongpt}
\bibfield{author}{\bibinfo{person}{Zongxin Yang}, \bibinfo{person}{Guikun Chen}, \bibinfo{person}{Xiaodi Li}, \bibinfo{person}{Wenguan Wang}, {and} \bibinfo{person}{Yi Yang}.} \bibinfo{year}{2024}\natexlab{a}.
\newblock \showarticletitle{Doraemongpt: Toward understanding dynamic scenes with large language models (exemplified as a video agent)}. In \bibinfo{booktitle}{\emph{ICML}}.
\newblock


\bibitem[Yue et~al\mbox{.}(2019)]%
        {yue2019domain}
\bibfield{author}{\bibinfo{person}{Xiangyu Yue}, \bibinfo{person}{Yang Zhang}, \bibinfo{person}{Sicheng Zhao}, \bibinfo{person}{Alberto Sangiovanni-Vincentelli}, \bibinfo{person}{Kurt Keutzer}, {and} \bibinfo{person}{Boqing Gong}.} \bibinfo{year}{2019}\natexlab{}.
\newblock \showarticletitle{Domain randomization and pyramid consistency: Simulation-to-real generalization without accessing target domain data}. In \bibinfo{booktitle}{\emph{ICCV}}.
\newblock


\bibitem[Zhang et~al\mbox{.}(2021)]%
        {PanZhang2021PrototypicalPL}
\bibfield{author}{\bibinfo{person}{Pan Zhang}, \bibinfo{person}{Bo Zhang}, \bibinfo{person}{Ting Zhang}, \bibinfo{person}{Dong Chen}, \bibinfo{person}{Yong Wang}, {and} \bibinfo{person}{Fang Wen}.} \bibinfo{year}{2021}\natexlab{}.
\newblock \showarticletitle{Prototypical pseudo label denoising and target structure learning for domain adaptive semantic segmentation}. In \bibinfo{booktitle}{\emph{CVPR}}.
\newblock


\bibitem[Zhang et~al\mbox{.}(2019b)]%
        {zhang2019category}
\bibfield{author}{\bibinfo{person}{Qiming Zhang}, \bibinfo{person}{Jing Zhang}, \bibinfo{person}{Wei Liu}, {and} \bibinfo{person}{Dacheng Tao}.} \bibinfo{year}{2019}\natexlab{b}.
\newblock \showarticletitle{Category anchor-guided unsupervised domain adaptation for semantic segmentation}. In \bibinfo{booktitle}{\emph{NeurIPS}}.
\newblock


\bibitem[Zhang et~al\mbox{.}(2022)]%
        {zhang2022implicit}
\bibfield{author}{\bibinfo{person}{Xinyu Zhang}, \bibinfo{person}{Dongdong Li}, \bibinfo{person}{Zhigang Wang}, \bibinfo{person}{Jian Wang}, \bibinfo{person}{Errui Ding}, \bibinfo{person}{Javen~Qinfeng Shi}, \bibinfo{person}{Zhaoxiang Zhang}, {and} \bibinfo{person}{Jingdong Wang}.} \bibinfo{year}{2022}\natexlab{}.
\newblock \showarticletitle{Implicit sample extension for unsupervised person re-identification}. In \bibinfo{booktitle}{\emph{CVPR}}.
\newblock


\bibitem[Zhang et~al\mbox{.}(2024)]%
        {zhang2024self}
\bibfield{author}{\bibinfo{person}{Xuanmeng Zhang}, \bibinfo{person}{Zhedong Zheng}, \bibinfo{person}{Minyue Jiang}, {and} \bibinfo{person}{Xiaoqing Ye}.} \bibinfo{year}{2024}\natexlab{}.
\newblock \showarticletitle{Self-ensembling depth completion via density-aware consistency}.
\newblock \bibinfo{journal}{\emph{Pattern Recognition}}  \bibinfo{volume}{154} (\bibinfo{year}{2024}), \bibinfo{pages}{110618}.
\newblock


\bibitem[Zhang et~al\mbox{.}(2023)]%
        {zhang2023boosting}
\bibfield{author}{\bibinfo{person}{Yurong Zhang}, \bibinfo{person}{Liulei Li}, \bibinfo{person}{Wenguan Wang}, \bibinfo{person}{Rong Xie}, \bibinfo{person}{Li Song}, {and} \bibinfo{person}{Wenjun Zhang}.} \bibinfo{year}{2023}\natexlab{}.
\newblock \showarticletitle{Boosting video object segmentation via space-time correspondence learning}. In \bibinfo{booktitle}{\emph{CVPR}}.
\newblock


\bibitem[Zhang et~al\mbox{.}(2018)]%
        {zhang2018joint}
\bibfield{author}{\bibinfo{person}{Zhenyu Zhang}, \bibinfo{person}{Zhen Cui}, \bibinfo{person}{Chunyan Xu}, \bibinfo{person}{Zequn Jie}, \bibinfo{person}{Xiang Li}, {and} \bibinfo{person}{Jian Yang}.} \bibinfo{year}{2018}\natexlab{}.
\newblock \showarticletitle{Joint task-recursive learning for semantic segmentation and depth estimation}. In \bibinfo{booktitle}{\emph{ECCV}}.
\newblock


\bibitem[Zhang et~al\mbox{.}(2019a)]%
        {zhang2019pattern}
\bibfield{author}{\bibinfo{person}{Zhenyu Zhang}, \bibinfo{person}{Zhen Cui}, \bibinfo{person}{Chunyan Xu}, \bibinfo{person}{Yan Yan}, \bibinfo{person}{Nicu Sebe}, {and} \bibinfo{person}{Jian Yang}.} \bibinfo{year}{2019}\natexlab{a}.
\newblock \showarticletitle{Pattern-affinitive propagation across depth, surface normal and semantic segmentation}. In \bibinfo{booktitle}{\emph{CVPR}}.
\newblock


\bibitem[Zhao et~al\mbox{.}(2022)]%
        {zhao2022source}
\bibfield{author}{\bibinfo{person}{Yuyang Zhao}, \bibinfo{person}{Zhun Zhong}, \bibinfo{person}{Zhiming Luo}, \bibinfo{person}{Gim~Hee Lee}, {and} \bibinfo{person}{Nicu Sebe}.} \bibinfo{year}{2022}\natexlab{}.
\newblock \showarticletitle{Source-free open compound domain adaptation in semantic segmentation}.
\newblock \bibinfo{journal}{\emph{IEEE Transactions on Circuits and Systems for Video Technology}} \bibinfo{volume}{32}, \bibinfo{number}{10} (\bibinfo{year}{2022}), \bibinfo{pages}{7019--7032}.
\newblock


\bibitem[Zheng and Yang(2020)]%
        {zheng2019unsupervised}
\bibfield{author}{\bibinfo{person}{Zhedong Zheng} {and} \bibinfo{person}{Yi Yang}.} \bibinfo{year}{2020}\natexlab{}.
\newblock \showarticletitle{Unsupervised Scene Adaptation with Memory Regularization in vivo}. In \bibinfo{booktitle}{\emph{IJCAI}}.
\newblock


\bibitem[Zheng and Yang(2021)]%
        {zheng2021rectifying}
\bibfield{author}{\bibinfo{person}{Zhedong Zheng} {and} \bibinfo{person}{Yi Yang}.} \bibinfo{year}{2021}\natexlab{}.
\newblock \showarticletitle{Rectifying pseudo label learning via uncertainty estimation for domain adaptive semantic segmentation}.
\newblock \bibinfo{journal}{\emph{International Journal of Computer Vision}} \bibinfo{volume}{129}, \bibinfo{number}{4} (\bibinfo{year}{2021}), \bibinfo{pages}{1106--1120}.
\newblock


\bibitem[Zheng and Yang(2022)]%
        {zheng2022adaptive}
\bibfield{author}{\bibinfo{person}{Zhedong Zheng} {and} \bibinfo{person}{Yi Yang}.} \bibinfo{year}{2022}\natexlab{}.
\newblock \showarticletitle{Adaptive Boosting for Domain Adaptation: Toward Robust Predictions in Scene Segmentation}.
\newblock \bibinfo{journal}{\emph{IEEE Transactions on Image Processing}}  \bibinfo{volume}{31} (\bibinfo{year}{2022}), \bibinfo{pages}{5371--5382}.
\newblock
\urldef\tempurl%
\url{https://doi.org/10.1109/TIP.2022.3195642}
\showDOI{\tempurl}


\bibitem[Zhou et~al\mbox{.}(2019)]%
        {zhou2019semantic}
\bibfield{author}{\bibinfo{person}{Bolei Zhou}, \bibinfo{person}{Hang Zhao}, \bibinfo{person}{Xavier Puig}, \bibinfo{person}{Tete Xiao}, \bibinfo{person}{Sanja Fidler}, \bibinfo{person}{Adela Barriuso}, {and} \bibinfo{person}{Antonio Torralba}.} \bibinfo{year}{2019}\natexlab{}.
\newblock \showarticletitle{Semantic understanding of scenes through the ade20k dataset}.
\newblock \bibinfo{journal}{\emph{International Journal of Computer Vision}}  \bibinfo{volume}{127} (\bibinfo{year}{2019}), \bibinfo{pages}{302--321}.
\newblock


\bibitem[Zhou et~al\mbox{.}(2022)]%
        {zhou2022context}
\bibfield{author}{\bibinfo{person}{Qianyu Zhou}, \bibinfo{person}{Zhengyang Feng}, \bibinfo{person}{Qiqi Gu}, \bibinfo{person}{Jiangmiao Pang}, \bibinfo{person}{Guangliang Cheng}, \bibinfo{person}{Xuequan Lu}, \bibinfo{person}{Jianping Shi}, {and} \bibinfo{person}{Lizhuang Ma}.} \bibinfo{year}{2022}\natexlab{}.
\newblock \showarticletitle{Context-aware mixup for domain adaptive semantic segmentation}.
\newblock \bibinfo{journal}{\emph{IEEE Transactions on Circuits and Systems for Video Technology}} (\bibinfo{year}{2022}).
\newblock


\bibitem[Zhou and Wang(2024)]%
        {zhou2024prototype}
\bibfield{author}{\bibinfo{person}{Tianfei Zhou} {and} \bibinfo{person}{Wenguan Wang}.} \bibinfo{year}{2024}\natexlab{}.
\newblock \showarticletitle{Prototype-based semantic segmentation}.
\newblock \bibinfo{journal}{\emph{IEEE Transactions on Pattern Analysis and Machine Intelligence}} (\bibinfo{year}{2024}).
\newblock


\bibitem[Zhou et~al\mbox{.}(2020b)]%
        {zhou2020target}
\bibfield{author}{\bibinfo{person}{Tianfei Zhou}, \bibinfo{person}{Wenguan Wang}, \bibinfo{person}{Yazhou Yao}, {and} \bibinfo{person}{Jianbing Shen}.} \bibinfo{year}{2020}\natexlab{b}.
\newblock \showarticletitle{Target-aware adaptive tracking for unsupervised video object segmentation}. In \bibinfo{booktitle}{\emph{CVPR Workshop}}.
\newblock


\bibitem[Zhou et~al\mbox{.}(2020a)]%
        {zhou2020affinity}
\bibfield{author}{\bibinfo{person}{Wei Zhou}, \bibinfo{person}{Yukang Wang}, \bibinfo{person}{Jiajia Chu}, \bibinfo{person}{Jiehua Yang}, \bibinfo{person}{Xiang Bai}, {and} \bibinfo{person}{Yongchao Xu}.} \bibinfo{year}{2020}\natexlab{a}.
\newblock \showarticletitle{Affinity space adaptation for semantic segmentation across domains}.
\newblock \bibinfo{journal}{\emph{IEEE Transactions on Image Processing}}  \bibinfo{volume}{30} (\bibinfo{year}{2020}), \bibinfo{pages}{2549--2561}.
\newblock


\bibitem[Zou et~al\mbox{.}(2018)]%
        {YangZou2018UnsupervisedDA}
\bibfield{author}{\bibinfo{person}{Yang Zou}, \bibinfo{person}{Zhiding Yu}, \bibinfo{person}{BVK Kumar}, {and} \bibinfo{person}{Jinsong Wang}.} \bibinfo{year}{2018}\natexlab{}.
\newblock \showarticletitle{Unsupervised domain adaptation for semantic segmentation via class-balanced self-training}. In \bibinfo{booktitle}{\emph{ECCV}}.
\newblock


\bibitem[Zou et~al\mbox{.}(2019)]%
        {zou2019confidence}
\bibfield{author}{\bibinfo{person}{Yang Zou}, \bibinfo{person}{Zhiding Yu}, \bibinfo{person}{Xiaofeng Liu}, \bibinfo{person}{BVK Kumar}, {and} \bibinfo{person}{Jinsong Wang}.} \bibinfo{year}{2019}\natexlab{}.
\newblock \showarticletitle{Confidence regularized self-training}. In \bibinfo{booktitle}{\emph{ICCV}}.
\newblock


\end{thebibliography}
}


\end{document}